\documentclass[10pt,twocolumn,letterpaper]{article}

\usepackage[]{cvpr}      % To produce the REVIEW version
\usepackage{nicefrac} 
\usepackage{colortbl}
\usepackage{graphicx}
\usepackage{subcaption}
\usepackage{xspace}
\usepackage{multirow}
\usepackage{makecell}
\usepackage{algorithm}
\usepackage{xcolor}
\newcommand{\bc}{\mathbf{c}}

\newcommand{\bJ}{\mathbf{J}}

\newcommand{\bp}{\mathbf{p}}

\newcommand{\bR}{\mathbf{R}}
\newcommand{\bS}{\mathbf{S}}

\newcommand{\bW}{\mathbf{W}}
\newcommand{\bx}{\mathbf{x}}

\newcommand{\bSigma}{\boldsymbol{\Sigma}}

\newcommand{\cG}{\mathcal{G}}

\makeatletter
\DeclareRobustCommand\onedot{\futurelet\@let@token\@onedot}
\def\@onedot{\ifx\@let@token.\else.\null\fi\xspace}

\def\etal{et~al\onedot}

\makeatother

\definecolor{yellow}{rgb}{1, 1, 0.7}
\definecolor{orange}{rgb}{1, 0.85, 0.7}
\definecolor{dark_orange}{rgb}{0.99215686, 0.60784314, 0.06666667}
\definecolor{tablered}{rgb}{1, 0.7, 0.7}
\definecolor{red}{rgb}{1, 0, 0}
\definecolor{wincolor}{rgb}{0.85, 0.0, 0.0}
\definecolor{darkyellow}{rgb}{0.8, 0.8, 0.5}
\definecolor{darkred}{rgb}{0.7, 0.3, 0.3}
\definecolor{darkgreen}{rgb}{0.3, 0.7, 0.3}
\definecolor{green}{rgb}{0, 1.0, 0}
\definecolor{pink}{rgb}{1, 0.4, 0.7}

\newcommand{\scenename}[1]{\textit{#1}}
\newcommand{\myparagraph}[1]{\vspace{0.5em} \noindent {\bf #1}\,\,\,}

\definecolor{iccvblue}{rgb}{0.21,0.49,0.74}
\usepackage[pagebackref,breaklinks,colorlinks,citecolor=iccvblue]{hyperref}

\title{LOD-GS: Level-of-Detail-Sensitive 3D Gaussian Splatting for \\Detail Conserved Anti-Aliasing}

\author{Zhenya Yang\textsuperscript{\rm 1}~~~Bingchen Gong\textsuperscript{\rm 2}~~~Kai Chen\textsuperscript{\rm 1} \\
\textsuperscript{\rm 1} The Chinese University of Hong Kong 
\hspace{0.3cm} \textsuperscript{\rm 2} Ecole Polytechnique
}

\begin{document}

\maketitle
\begin{abstract}
Despite the advancements in quality and efficiency achieved by 3D Gaussian Splatting (3DGS) in 3D scene rendering, aliasing artifacts remain a persistent challenge.
Existing approaches primarily rely on low-pass filtering to mitigate aliasing.
However, these methods are not sensitive to the sampling rate, often resulting in under-filtering and over-smoothing renderings.
To address this limitation, we propose \textbf{LOD-GS}, a \textbf{L}evel-\textbf{o}f-\textbf{D}etail-sensitive filtering framework for \textbf{G}aussian \textbf{S}platting, which dynamically predicts the optimal filtering strength for each 3D Gaussian primitive. 
Specifically, we introduce a set of basis functions to each Gaussian, which take the sampling rate as input to model appearance variations, enabling sampling-rate-sensitive filtering. 
These basis function parameters are jointly optimized with the 3D Gaussian in an end-to-end manner.
The sampling rate is influenced by both focal length and camera distance.
However, existing methods and datasets rely solely on down-sampling to simulate focal length changes for anti-aliasing evaluation,
overlooking the impact of camera distance.
To enable a more comprehensive assessment, we introduce a new synthetic dataset featuring objects rendered at varying camera distances.
Extensive experiments on both public datasets and our newly collected dataset demonstrate that our method achieves SOTA rendering quality while effectively eliminating aliasing.
The code and dataset are available at \href{https://github.com/Huster-YZY/LOD-GS}{https://github.com/Huster-YZY/LOD-GS}.
\end{abstract}

%%%%%%%%% BODY TEXT

\vspace{-1em}
\section{Introduction}
\label{sec:introduction}
%general background of NVS
Novel view synthesis (NVS) plays a crucial role in the fields of computer vision and computer graphics. Advanced NVS techniques significantly enhance applications in virtual reality, digital modeling, and embodied AI. A notable milestone in NVS is the Neural Radiance Field (NeRF)~\cite{mildenhall2020nerf}, which represents a 3D scene using a multi-layer perceptron (MLP) and optimizes this MLP through volume ray marching~\cite{max1995optical,levoy1990efficient,brebin1998volume} and gradient descent based on multi-view inputs.
In comparison to NeRF~\cite{mildenhall2020nerf}, recent advancements in 3D Gaussian Splatting (3DGS)~\cite{kerbl3Dgaussians} offer an alternative approach by representing a 3D scene with a collection of 3D Gaussians. This method benefits from an efficiently implemented CUDA rasterizer, which facilitates real-time rendering and efficient training processes.
These advantages position 3DGS as a competitive alternative to NeRF.

%-----------------------------------------------------------------------------
\begin{figure}[!t] 
	\centering
    \includegraphics[width=\linewidth]{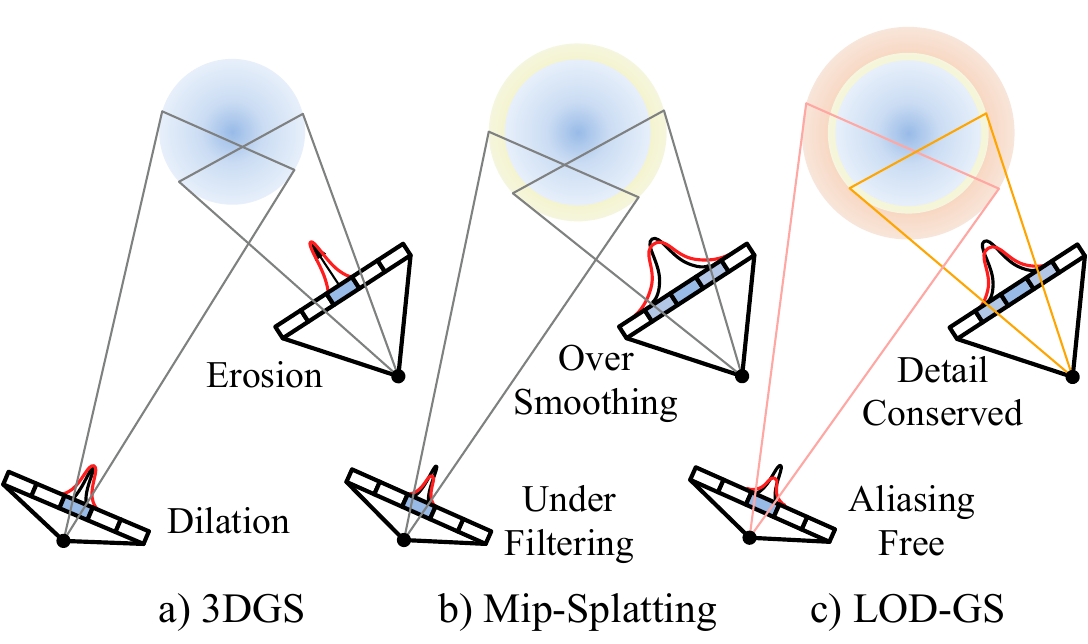}
	%	\vspace{-1mm}
	\caption{
    (a) 3DGS treats a 3D Gaussian primitive (in blue) uniformly across different views and applies a dilation operator (in red) before rendering.  
    (b) Mip-Splatting applies the fixed 3D smoothing filter (in yellow) to the primitive across all views.  
    (c) LOD-GS applies different filters (in yellow and orange) based on the sampling rates of the views.  
    Mip-Splatting utilizes 2D Mip filter (in red) and LOD-GS uses EWA filter (in red) before rendering.  
    The lack of filtering and the use of a dilation operator in 3DGS result in erosion and dilation effects. The fixed 3D smoothing filter in Mip-Splatting leads to over-smoothing and under-filtering.
}
	% \vspace{-1.5em}
	\label{fig:teaser}
\end{figure} 
%-----------------------------------------------------------------------------

%experiment observation
% single scale
% reasons
However, 3DGS also faces aliasing problems~\cite{crow1977aliasing} like NeRF.
% 3DGS is able to achieve good novel view synthesis performance when the training and testing views share roughly the same focal length and camera distance to the observed scene (i.e., same sampling rate of the camera).
3DGS is able to achieve good novel view synthesis performance when the training and testing views share roughly the same sampling rate to the observed scene, defined as the ratio of focal length to camera distance.
So the aliasing problem of 3DGS is unobvious because existing datasets usually capture a scene from nearly the same distance with the same focal length. 
For 3DGS models trained on these single-scale datasets, they tend to generate degraded results or aliased renderings when test views zoom in or out significantly. 
The main reason for this problem is that the 3DGS model is fixed after training.
It does not adjust its appearance according to the sampling rate.
This limitation leads to the aliasing and degraded results.

% inspiration-lod-mipmap
In this paper, we are inspired by the Mipmap technique~\cite{mipmap}, which is used in computer graphics rendering pipelines to tackle the aliasing.
Mipmap involves a collection of progressively downsampled textures,  enabling the program to choose the appropriate resolution based on the camera's sampling rate——a technique known as pre-filtering.
This method is grounded in the Nyquist-Shannon Sampling Theorem~\cite{nyquist2009certain,shannon1949communication} and its concept can be generalized as Level of Detail (LOD)~\cite{lod-of-3d,weiler2000level,lamar1999multiresolution,de1995hierarchical}, which indicates that the appearance of an object changes with the sampling rate for the scene.

\begin{figure*}[!t] 
	\centering
    \includegraphics[width=\linewidth]{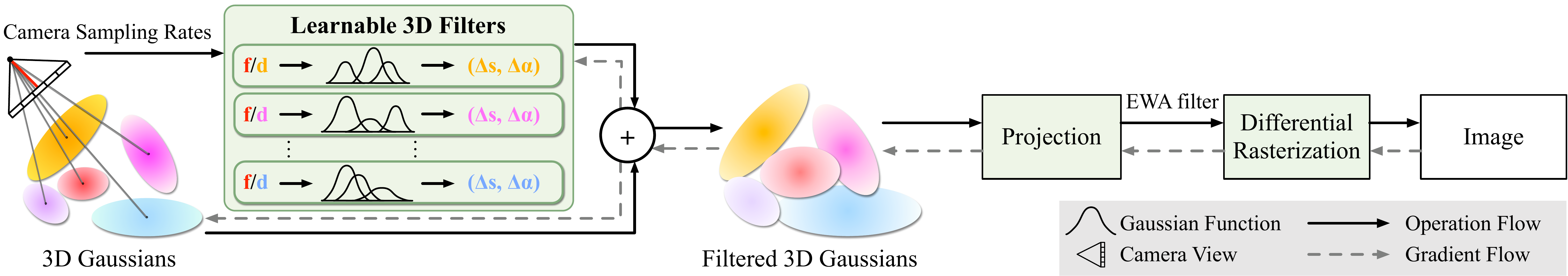}
	% \vspace{1mm}
	\caption{
    \textbf{Overview of our pipeline.}
    During rendering, the sampling rate for each Gaussian primitive is sent to a learnable module that predicts the appropriate 3D filters for them. The filtered 3D Gaussians are then projected into 2D space and further processed using EWA filter before rasterization. The learnable 3D filters and the 3D Gaussians are jointly optimized end-to-end with image supervision.
    }
	% \vspace{-1.5em}
	\label{fig:overview}
\end{figure*}

% directly multi-scale training --> fig1 analysis
% mip-nerf
%our method(lod filtering + ewa filtering)-->distangle
% 3dgs aliasing methods comparison
We adopt the Mipmap technique into the 3D Gaussian Splatting framework~\cite{kerbl3Dgaussians} to represent the pre-filtered radiance field with different levels of detail using multiscale images as input.
Because the vanilla 3DGS is not sensitive to the sampling rate, training it with multi-scale images can lead to ambiguity in optimization, making rendering results blur and lack of detail.
In this paper, we propose a level-of-detail-sensitive 3DGS framework, named LOD-GS, to effectively sense the change in sampling rate and train the pre-filtered radiance field from pre-filtered images.
We propose to add a set of basis functions on each Gaussian primitive to learn the appearance change across different sampling rates.
These basis functions take the sampling rate as input and predict the filter size and appearance change for each primitive as illustrated in Figure~\ref{fig:teaser} and Figure~\ref{fig:overview}.
The filtered primitive will be splatted into 2D screen space and perform the Elliptical Weighted Average (EWA) filtering~\cite{zwicker2001ewa} to further improve anti-aliasing ability of our method.
Besides the framework, we re-render the synthetic dataset used in NeRF~\cite{mildenhall2020nerf} from three different camera distances to simulate the changes in sampling rate caused by varying camera distances.
The evaluation is carried out on both public datasets~\cite{mildenhall2020nerf, mip-nerf, mipnerf360} and our newly collected dataset.
Experiment results demonstrate that our method can disentangle training views with different sampling rates and learn a pre-filtered radiance field from these views, achieving detail-conserved and aliasing-free rendering at the same time.

Our contributions are as follows:
\begin{itemize}
    \item 
    We introduce \textbf{LOD-GS}, a Level-of-Detail-sensitive Gaussian Splatting framework that enables effective anti-aliasing while preserving fine details in rendering results.
    \item 
    Our proposed LOD-GS eliminates the ambiguity in training caused by inputs with different sampling rates, allowing it to effectively learn a pre-filtered radiance field from pre-filtered images.
    \item 
    We extend the original NeRF Synthetic Dataset~\cite{mildenhall2020nerf} by incorporating rendering views from different camera distances, providing a more comprehensive evaluation of zoom-in and zoom-out effects in neural radiance fields.
\end{itemize}

%-----------------------------------------------------------------------------
\begin{figure}[!t] 
	\centering
    \includegraphics[width=\linewidth]{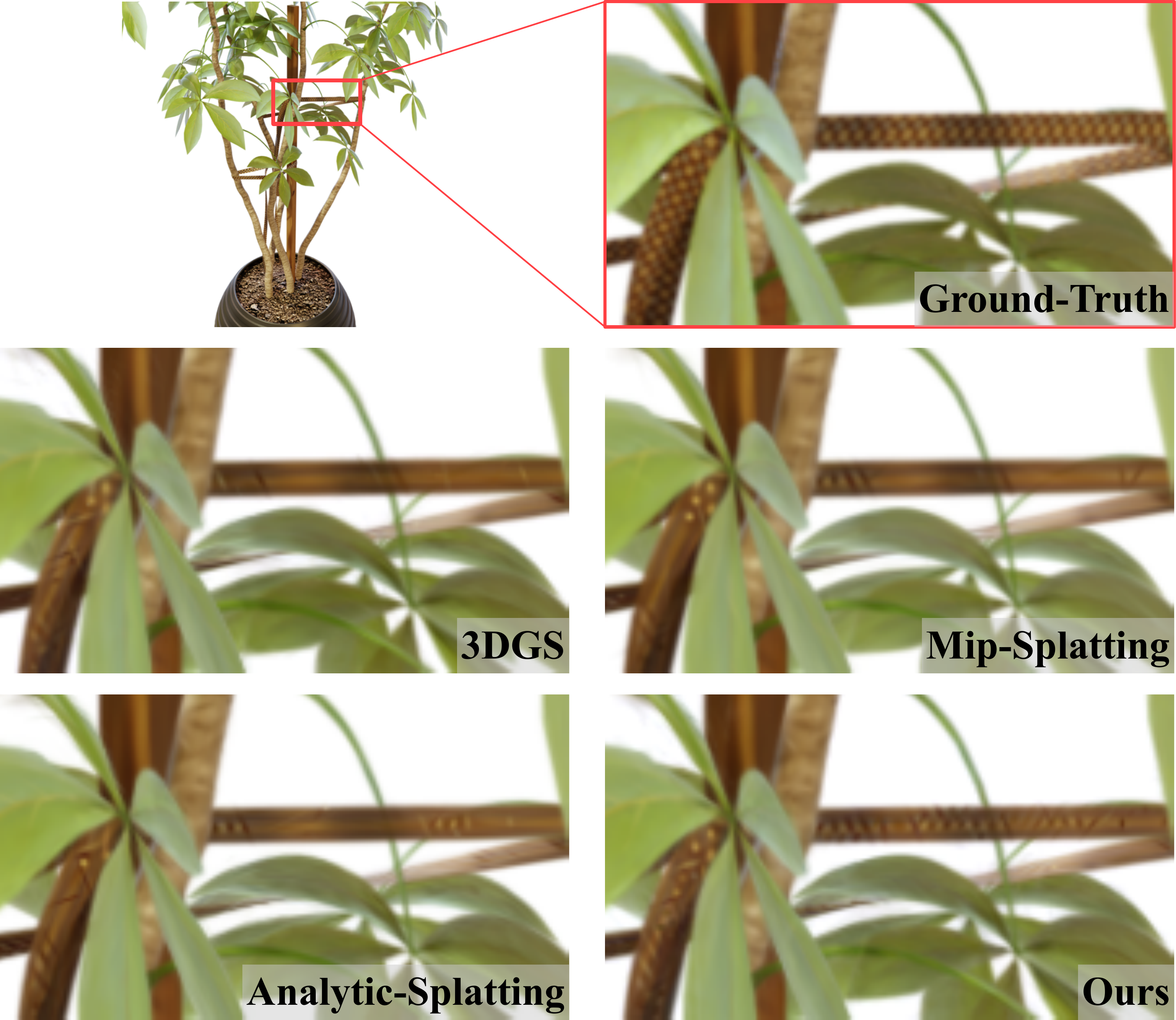}
	%	\vspace{-1mm}
	\caption{
    \textbf{Comparison of Detail Reconstruction.} All methods are trained on inputs with varying sampling rates. In comparing the reconstruction results of different methods, only our LOD-GS successfully reconstructs the intricate texture of the silk ribbon.
    Other methods share the smoothing problem to different extents.
}
	% \vspace{-1.5em}
	\label{fig:simple-comparison}
\end{figure} 
%-----------------------------------------------------------------------------

%-----------------------------------------------------------------------------
\section{Related Work}
\label{sec:relatedWork}
%-----------------------------------------------------------------------------
%Our work follows the overall framework of neural radiance field (NeRF)~\cite{mildenhall2020nerf}, and is devoted to reducing the aliasing and blurring artifacts while improving the efficiency in both training and rendering time.
%%
%Here we first review the anti-aliasing techniques in the computer graphics literature and then the works tailor-made for reducing computation cost for NeRF.

%-----------------------------------------------------------------------------
% \TODO{Restructure it to be more NeRF-related.}
% \vspace{-4mm}
\myparagraph{Novel View Synthesis}
Novel View Synthesis (NVS) is a fundamental vision task aimed at generating plausible renderings from new camera positions based on a set of input images and their corresponding camera positions.
The introduction of Neural Radiance Fields (NeRF)~\cite{mildenhall2020nerf} has significantly changed the approach to solving the NVS task.
NeRF formulates Novel View Synthesis as a volume rendering and optimization problem.
Nearly all subsequent works adhere to this framework.
However, due to the frequent querying of the MLP during training and inference, vanilla 
NeRF~\cite{mildenhall2020nerf} is quite slow, requiring over ten hours for training per scene and dozens of seconds to render a new view.
Following works have replaced the full implicit representation of NeRF with feature grid-based representations to accelerate training and inference speeds~\cite{chen2022tensorf,fridovich2022plenoxels,liu2020neural,sun2022direct,muller2022instant}.
Some approaches also explore the point-based neural representations~\cite{xu2022point,sun2024pointnerf++,lassner2021pulsar,prokudin2023dynamic,yifan2019differentiable,zheng2023pointavatar}.
The emergence of 3D Gaussian Splatting~\cite{kerbl3Dgaussians}, which represents the scene as a collection of 3D Gaussians, has removed the MLP from the radiance field representation. 
This specially designed MLP-free framework, combined with efficient implementation, allows 3DGS to achieve real-time rendering speeds and perform efficient training. Due to these advantages, 3DGS is beginning to replace NeRF in many applications. 
However, 3DGS is struggling to tackle the aliasing, this paper focuses on enhancing the anti-aliasing capability of 3DGS to achieve aliasing-free and detail conserved rendering.

\myparagraph{Anti-Aliasing of Neural Radiance Field}
Neural rendering methods integrate pre-filtering techniques to mitigate aliasing~\cite{zhuang2023lod,mip-nerf,mipnerf360,barron2023zip,trimip-nerf,mip-splatting}.
Mip-NeRF~\cite{mip-nerf} is the first work to address the aliasing problem in NeRF by proposing Integrated Positional Encoding (IPE), which filters out high-frequency components when the camera's sampling rate is low. 
Following works~\cite{trimip-nerf, mipnerf360} speed up the training of Mip-NeRF and extend it to handle unbounded scenes.
Due to the differences in rendering methods between NeRF and 3DGS, the anti-aliasing strategy used in NeRF cannot be directly applied to 3DGS. 
Several methods have recently been proposed to address the aliasing problem in Gaussian Splatting~\cite{mip-splatting,multiscale-3dgs,mipmapgs,analytic-splatting,sa-gs,octreegs}.
% Multiscale-3DGS (MSGS)~\cite{multiscale-3dgs} represents a 3D scene as a collection of 3D Gaussians of varying levels, selecting only a subset of these levels for rendering based on the target resolution to overcome aliasing.
Mip-Splatting~\cite{mip-splatting} introduces the 3D smoothing filter and the 2D mip-filter to remove high-frequency components from the Gaussians. 
Mipmap-GS~\cite{mipmapgs} utilizes a two stage training and generated pseudo ground truth to adapt to a specific resolution.
Analytic-Splatting~\cite{analytic-splatting} proposes an approximate method to compute the integration of Gaussian, effectively tackling the aliasing problem. 
However, all these methods have their limitations. 
% The explicit level representation in MSGS complicates the training and increase the number of primitives.
Mipmap-GS needs re-training once the sampling rate changes.
The introduction of integration in Analytic-Splatting slows down its training and rendering.
The 3D filter in Mip-Splatting remains fixed after training, regardless of the camera's sampling rate. Additionally, the 2D Mip filter cannot sense the existence of the 3D filter, potentially resulting in inappropriate 2D filtering degree. All these factors contribute to the problem of under-filtering and over-smoothing, as illustrated in Figure \ref{fig:simple-comparison}.
To solve these problems, we propose a filtering strategy which dynamically adjusts the filtering degree and learns this adjustment from data in an end-to-end manner.

\section{Preliminaries}
\label{sec:pre}
%\vspace{-2mm}
%-----------------------------------------------------------------------------

%-----------------------------------------------------------------------------
\subsection{3D Gaussian Splatting}
\label{subsec:overview}
Kerbl~\etal~\cite{kerbl3Dgaussians} utilize learnable 3D Gaussian primitives to represent 3D scenes and render different views using a differentiable volume splatting rasterizer. In 3DGS, each 3D Gaussian primitive is parameterized using a 3D covariance matrix $\bSigma$ and a distribution center $\bp_k$:
\begin{equation}
\cG(\bp) = \exp\left(-\frac{1}{2} (\bp - \bp_k)^\top \bSigma^{-1}(\bp - \bp_k)\right)
\label{eq:gaussian}
\end{equation}
During optimization, the covariance matrix $\bSigma$ is factorized into a scaling matrix $\bS$ and a rotation matrix $\bR$ as $\bSigma = \bR \bS \bS^\top \bR^\top$ to ensure its positive semidefiniteness.
To obtain the rendering results of 3D Gaussians from a specific view, the 3D Gaussian is first projected to a 2D splat in screen space using the view matrix $\bW$ and an affine approximated projection matrix $\bJ$ as illustrated in \cite{zwicker2001ewa}:
\begin{equation}
\bSigma' = \bJ \bW \bSigma \bW^\top \bJ^\top
\end{equation}
By removing the third row and column of $\bSigma'$, we obtain a $2 \times 2$ matrix, which represents the covariance matrix $\bSigma^{2D}$ of the 2D splat $\cG^{2D}$. Finally, the color of each pixel $\bx$ can be computed using volumetric alpha blending as follows:
\begin{equation}
\bc(\bx) = \sum_{k=1}^K \bc_k \alpha_k \cG^{2D}_k(\bx) \prod_{j=1}^{k-1} (1 - \alpha_j \cG^{2D}_j(\bx))
\end{equation}
where $k$ is the index of the Gaussian primitives covering the current pixel, $\alpha_k$ denotes the alpha values, and $\mathbf{c}_k$ represents the view-dependent appearance modeled using Spherical Harmonics. All attributes of the 3D Gaussian primitives $(\mathbf{p}, \mathbf{S}, \mathbf{R}, \alpha, \mathbf{c})$ are optimized using the photometric loss between the rendered images and the ground-truth images.

\subsection{Nyquist-Shannon Sampling Theorem}
% discrete characteristic --> sampling --> the problem of aliasing
% how Nyquist sampling theorem claim to avoid the aliasing
Many signals initially exist in analog form. Sampling converts these continuous signals into a discrete format suitable for processing, storage, and transmission by digital devices. However, improper sampling can cause distortion known as aliasing.
The \textbf{Nyquist-Shannon sampling theorem} states that \emph{the sampling rate $\nu$ must be at least twice the bandwidth of the signal to avoid aliasing.} 
Based on this theorem, there are two common approaches to address aliasing: one is to increase the sampling rate, which inevitably raises the computational workload; the other is to apply bandpass filters to the signals to eliminate frequencies higher than $\nicefrac{\nu}{2}$. The frequency $\nicefrac{\nu}{2}$ is known as the Nyquist frequency.

The rendering process of 3D Gaussian Splatting involves sampling. It uses discrete pixels to sample 3D Gaussians, ultimately producing a 2D rendering result from this 3D representation. Consequently, the rendering of 3D Gaussian Splatting also faces the challenge of aliasing.
Corresponding to the two approaches mentioned earlier, the first method to address aliasing in 3DGS rendering is Super Sampling~\cite{wang2022nerfsr-mm}. This technique renders images at a higher resolution—effectively increasing the sampling rate—and then pools the high-resolution results to generate anti-aliased low-resolution images. However, the high resolution used in Super Sampling leads to additional computational burden, resulting in slower rendering speeds.
The second approach applies filtering to 3D Gaussians before rendering, ensuring their frequency remains below the Nyquist limit.
We adopt this method for efficient rendering. However, excessive filtering can cause signal loss, leading to overly smooth results that lack fine details.
In Section~\ref{sec:method}, we will explain in detail how our method selects the appropriate filter for each Gaussian primitive.
% ###############################################
% -----------------------------------------------------
\renewcommand{\arraystretch}{1.1}
\begin{table*}[t]
    \renewcommand{\tabcolsep}{1pt}
    \centering
    \resizebox{1.0\linewidth}{!}{
    \begin{tabular}{@{}l@{\,\,}|ccccc|ccccc|ccccc}
    &   \multicolumn{5}{c|}{PSNR $\uparrow$} & \multicolumn{5}{c|}{SSIM $\uparrow$} & \multicolumn{5}{c}{LPIPS $\downarrow$}  \\
    & Full Res. & $\nicefrac{1}{2}$ Res. & $\nicefrac{1}{4}$ Res. & $\nicefrac{1}{8}$ Res. & Avg. & Full Res. & $\nicefrac{1}{2}$ Res. & $\nicefrac{1}{4}$ Res. & $\nicefrac{1}{8}$ Res. & Avg. & Full Res. & $\nicefrac{1}{2}$ Res. & $\nicefrac{1}{4}$ Res. & $\nicefrac{1}{8}$ Res & Avg.  \\ \hline
    NeRF~\cite{mildenhall2020nerf}&                  29.90 &                    32.13 &                    33.40 &                     29.47 &  31.23 &                   0.938 &                    0.959 &                    0.973 &                    0.962 &  0.958 &                  0.074 &                    0.040 &                    0.024 &                    0.039 &                  0.044 
\\
MipNeRF~\cite{mip-nerf}&  32.63 &     34.34 &35.47 &35.60 & 34.51 & 0.958 & 0.970 & 0.979 & 0.983 & 0.973& 0.047 & 0.026 & 0.017 & 0.012 & 0.026 
\\

Tri-MipRF~\cite{trimip-nerf}&   32.65 & 34.24 & 35.02 & 35.53 & 34.36 & 0.958 & 0.971 & 0.980 & 0.987 & 0.974 & 0.047 & 0.027 & 0.018 & 0.012 & 0.026
\\
\hline
3DGS~\cite{kerbl3Dgaussians}&  28.79 & 30.66 & 31.64 & 27.98 & 29.77 & 0.943 & 0.962 & 0.972 & 0.960 & 0.960 & 0.065 & 0.038 & 0.025 & 0.031 & 0.040
\\
Mipmap-GS~\cite{mipmapgs}&28.79 &30.67 &31.66 &28.00 & 29.78 &0.943 &0.962 &0.973 &0.961 &0.960 &0.065 &0.038 &0.025 &0.031 &0.040
\\
SA-GS~\cite{sa-gs}&30.80 &32.67 &35.06 &\cellcolor{yellow}35.77 &33.58 &0.956 &0.969 &0.980 &0.985 &0.973 &0.056 &0.032 &0.020 &0.014 &0.031
\\
Multiscale-3DGS\cite{multiscale-3dgs} &\cellcolor{tablered}33.36 &27.15&21.41&17.61&24.88 &\cellcolor{tablered}0.969 &0.951 &0.875 &0.764 &0.890 &\cellcolor{tablered}0.031 &0.032 &0.067 &0.126 &0.064
\\
Mip-Splatting~\cite{mip-splatting} &32.81 &\cellcolor{yellow}34.49 &\cellcolor{yellow}35.45 & 35.50 &\cellcolor{yellow}34.56 &\cellcolor{orange}0.967 &\cellcolor{tablered}0.977 &\cellcolor{yellow}0.983 &\cellcolor{yellow}0.988 &\cellcolor{orange}0.979 &\cellcolor{yellow}0.035 &\cellcolor{tablered}0.019 &\cellcolor{orange}0.013 &\cellcolor{orange}0.010 &\cellcolor{orange}0.019
\\
% Mipmap-GS~\cite{mipmapgs} &  - & - & - & - & - & - & - &- & - & - & - & - & - & - & -
% \\
Analytic-Splatting~\cite{analytic-splatting} &\cellcolor{orange}33.22 & \cellcolor{tablered}34.92 &\cellcolor{orange}35.98 & \cellcolor{orange}36.00 &\cellcolor{orange}35.03 &\cellcolor{orange}0.967 &\cellcolor{tablered}0.977 &\cellcolor{orange}0.984 &\cellcolor{orange}0.989 &\cellcolor{orange}0.979 &\cellcolor{orange}0.033 &\cellcolor{tablered}0.019 & \cellcolor{tablered}0.012 &\cellcolor{orange}0.010 &\cellcolor{tablered}0.018 
\\

\hline
LOD-GS (ours) &\cellcolor{yellow}32.90 &\cellcolor{orange}34.88 &\cellcolor{tablered}36.43 &\cellcolor{tablered}37.27 &\cellcolor{tablered}35.37 &\cellcolor{yellow}0.966 &\cellcolor{tablered}0.977 &\cellcolor{tablered}0.985 &\cellcolor{tablered}0.990 &\cellcolor{tablered}0.980 &\cellcolor{yellow}0.035 &\cellcolor{tablered}0.019 &\cellcolor{tablered}0.012 &\cellcolor{tablered}0.008 &\cellcolor{tablered} 0.018

% \\
% \hline
% w/o LOD filter &31.88 &34.09 &35.86 &35.73 &34.39 &0.961 &0.975 &0.984 &0.989 &0.977 &0.042 &0.021 &0.013 &0.010 &0.022
% \\
% w/o EWA filter &32.63 &34.39 &35.66 &35.76 &34.61 &0.966 &0.977 &0.983 &0.987 &0.978 &0.038 &0.020 &0.013 &0.010 &0.020
    \end{tabular}
    }
    % \vspace{0.1in}
    \caption{
    \textbf{Multi-scale Training and Multi-scale Testing on the Blender dataset~\cite{mildenhall2020nerf}.} Our approach demonstrates state-of-the-art performance across most metrics and is highly competitive with existing GS-based methods specifically designed for anti-aliasing, such as Mipmap-GS~\cite{mipmapgs}, SA-GS~\cite{sa-gs}, Multiscale-3DGS~\cite{multiscale-3dgs}, Mip-Splatting~\cite{mip-splatting},  and Analytic-Splatting~\cite{analytic-splatting}.
    }
    \label{tab:avg_multiblender_results}
    \vspace{-0.1in}
\end{table*}
% -----------------------------------------------------

% \begin{figure}[!t] 
% 	\centering
%     \includegraphics[width=\linewidth]{fig/sampling.pdf}
%         \vspace{0.5em}
% 	\caption{
%     \textbf{Sampling Interval of Camera.} We give an intuitive explanation of sampling interval, which is actually the footprint of one pixel in 3D scene. The result is computed through approximate calculations using radians.
%     }
% 	% \vspace{-1.5em}
% 	\label{fig:sampling}
% \end{figure} 

\section{Method}
\label{sec:method}
In this part, we first introduce our Level-of-Detail-Sensitive 3D Filter in Section~\ref{subsec:lod_implementation}, which takes the sampling rate as input and predicts the filters for 3D Gaussians. 
This design helps 3DGS disentangle the inputs of different sampling rates to better learn a pre-filtered radiance field from per-filtered images without ambiguity.
Then we illustrate the EWA filter technique in Section~\ref{subsec:ewa}, which is employed to enhance the anti-aliasing capabilities of our method.
The overview of our method is illustrated in Figure~\ref{fig:overview}.

\subsection{Level-of-Detail-Sensitive 3D Filter}
\label{subsec:lod_implementation}
Applying zoom-in and zoom-out operations on scenes reconstructed by 3DGS can lead to noticeable visual artifacts, including erosion and dilation~\cite{mip-splatting}.
To solve these artifacts, Mip-Splatting proposes 3D Smoothing Filter as follows :
\begin{equation}
    \mathcal{G}_k(\mathrm{x}) = \sqrt{\frac{|\Sigma_k|}{|\Sigma_k + \frac{s}{\hat{\nu_k}}|}} e^{-\frac{1}{2}(\mathrm{x}-\mathbf{p}_k)^T(\Sigma_{k} + \frac{s}{\hat{\nu_k}})(\mathrm{x}-\mathbf{p}_k)}
    \label{eq:mip}
\end{equation}
where the hyper parameter $s$ is used to control the filter size and $\hat{\nu_k}$ is the maximal sampling rate for the $k$-th primitive which could be computed as follows:
\begin{equation}
    \hat{\nu_k} = \max_{n \in \{1,\ldots, N\}}\frac{f_n}{d_n}
    \label{eq:mip-freq}
\end{equation}
\vspace{-1pt}
where \( N \) is the number of training views, \( f_n \) is the focal length of the \( n \)-th view, and \( d_n \) is the distance from the \( n \)-th view to the current primitive.
The time complexity of Equation~\ref{eq:mip-freq} is \(\mathcal{O}(KN)\), where \(K\) is the number of Gaussian primitives and \(N\) is the number of training views.
To reduce computational load, Mip-Splatting recomputes \( \hat{\nu_k} \) every 100 iterations. After training, the \( \hat{\nu} \) for each primitive is fixed, meaning that during testing, the filter size remains unchanged regardless of variations in the sampling rate. 
As a result, the training process becomes dependent on the choice of the hyperparameter \( s \). 
As shown in Figure~\ref{fig:teaser}, Figure~\ref{fig:simple-comparison} and Figure~\ref{fig:filter-cmp}, a fixed 3D smoothing filter can lead to some textures being over-filtered while others are not sufficiently filtered when the test camera moves closer or farther.

In the \emph{Level of Detail (LOD) concept}~\cite{lod-of-3d,weiler2000level,lamar1999multiresolution,de1995hierarchical}, an object’s texture resolution adapts based on the sampling rate. Inspired by this idea, we design a learnable framework that takes the sampling rates of Gaussian primitives as input and outputs appropriate 3D filters.
In other words, the filter size of each primitive dynamically adjusts according to the sampling rate. 
To minimize the increase in computational workload of training and inference, we design a learnable Gaussian Mixture Model (GMM) module instead of using a MLP for each primitive to predict the suitable filter size.
The added GMM module can be expressed using the following equation:
\begin{equation}
    \mathcal{F}(x) = \sum_{i=1}^{l}w_{i}\exp(-\frac{(x-\mu_i)^2}{2\sigma_{i}^2})
\end{equation}
\vspace{-1pt}
where $l$ is the number of basis functions, $u_i$ and $\sigma_i$ are the distribution center and standard deviation which will be optimized using gradient descent.
The input of the GMM module is the sampling rate, defined as: $\nu = \nicefrac{1}{T} =  \nicefrac{f}{d}$,
where $f$ is the focal length and $d$ is the distance from camera to the primitive.
$T$ is defined as the sampling interval.
We also add a learnable residual in opacity to model the opacity change during the filtering, this design helps us avoid the use of hyperparameter in Equation~\ref{eq:mip}.
Above all, our proposed LOD sensitive filter could be represented as:
\begin{equation}
    \mathcal{G}_k(\mathrm{x}) = e^{-\frac{1}{2}(\mathrm{x}-\mathbf{p}_k)^T(\Sigma_{k} + \mathcal{F}_{s}(\nu))(\mathrm{x}-\mathbf{p}_k)}
\end{equation}
\vspace{-1pt}
where \( \nu \) is the sampling rate of the Gaussian primitive. 
The sampling rate from one camera to all the primitives can be computed efficiently in \(\mathcal{O}(K)\). This allows us to perform this computation in each iteration, rather than updating the sampling rate every 100 iterations, as done in Mip-Splatting~\cite{mip-splatting}. 
This design makes each Gaussian primitive sensitive to changes in the sampling rate. As a result, the scale and opacity of the primitive can be adjusted accordingly. 
The following experimental results demonstrate that our proposed method can better bake images from different sampling rates into a single radiance field, generating aliasing-free rendering results while preserving details.

\subsection{EWA Filtering}
\label{subsec:ewa}
In the implementation of 3DGS, a dilation operator~\cite{mip-splatting} is applied to each projected Gaussian. This helps avoid small primitives that are hard to optimize.
The dilation operator could be written as: $\mathcal{G}_k^{2D}(\mathrm{x}) = e^{-\frac{1}{2}(\mathrm{x}-\mathbf{p}_k)^T(\Sigma_k^{2D}+s\mathbf{I})(\mathrm{x}-\mathbf{p}_k)}$.
This operator leads to obvious dilation artifacts when the image resolution decreases. 
These artifacts are primarily caused by the expansion of the primitive's scale without adjusting its opacity. As a result, the overall energy increases, leading to the dilation effect.
The EWA Filtering~\cite{zwicker2001ewa} could be used to alleviate this artifact:
\begin{equation}
    \mathcal{G}_k^{2D}(\mathrm{x}) = \sqrt{\frac{|\Sigma_k^{2D}|}{|\Sigma_k^{2D} + s\mathbf{I}|}} e^{-\frac{1}{2}(\mathrm{x}-\mathbf{p}_k)^T(\Sigma_k^{2D}+s\mathbf{I})(\mathrm{x}-\mathbf{p}_k)}
\end{equation}
% \vspace{-3pt} 
The main difference between dilation operator and EWA filter is the normalization term preceding the exponential term.
In LOD-GS, we use the EWA filter before the alpha blending computation to reduce the dilation and aliasing artifact.
After the EWA filtering, we perform alpha blending to obtain the rendering result as follows:
\begin{align}
    \bc(\bx) &= \sum_{k=1}^K  \bc_k  \hat{\alpha_k} \mathcal{G}_k^{2D}(\mathrm{x})\prod_{j=1}^{k-1} (1 - \hat{\alpha_j}\mathcal{G}_j^{2D}(\mathrm{x}))\\
    \hat{\alpha}_k  &=\alpha_k + \mathcal{F}_{\alpha}(\nu)
\end{align}
\vspace{-1pt}
where \( \hat{\alpha} \) is the opacity of the LOD-filtered Gaussian primitive. 
After obtaining the rendering results, we compute the image loss between these results and the ground truth. 
Then we perform gradient descent to optimize all Gaussian parameters and the filter prediction module in an end-to-end manner.
This design enables the LOD filter to adjust its filtering degree by taking the EWA filter into account, allowing it to predict the most appropriate filter for achieving both aliasing-free and detail-conserved rendering.
Conversely, the 3D filter and 2D Mip filter of Mip-Splatting operate sequentially but independently, which can easily result in over-filtering and lead to blurry rendering results.

\begin{figure*}[!t] 
	\centering
    \includegraphics[width=\linewidth]{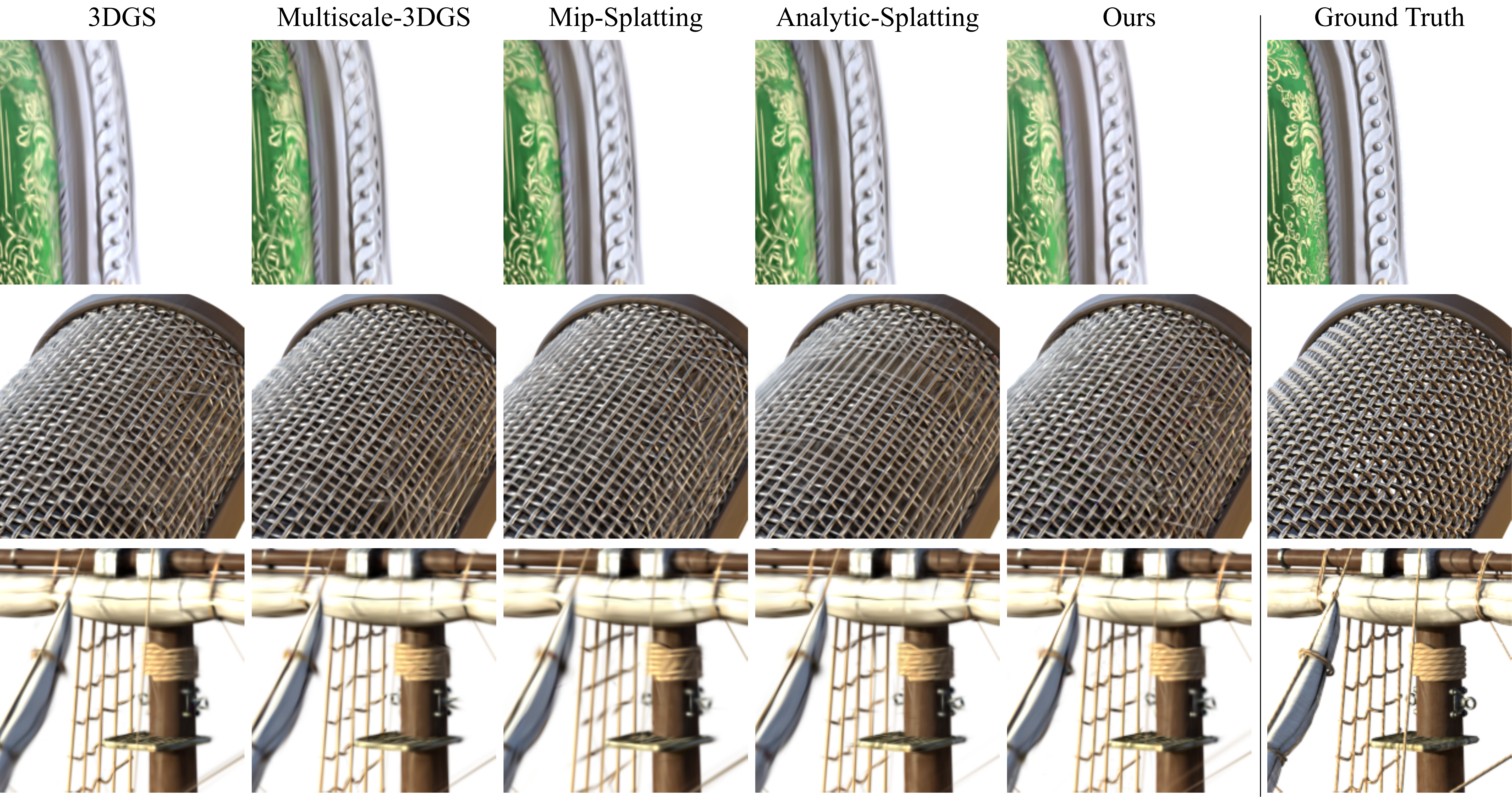}
	% \vspace{1mm}
    \vspace{-1.5em}
	\caption{
    \textbf{Qualitative comparison on our extended Blender dataset.} All methods are trained across three levels: \(L_1\) (near), \(L_2\) (middle), and \(L_3\) (far). Our method more effectively captures the details of microstructures and textures during the multi-level training.
    }
	% \vspace{-0.5em}
	\label{fig:detail-cmp}
\end{figure*} 

\renewcommand{\arraystretch}{1.1}
\begin{table*}[]
    \renewcommand{\tabcolsep}{3pt}
    \centering
    \resizebox{0.95\linewidth}{!}{
    \begin{tabular}{@{}l@{\,\,}|cccc|cccc|cccc}
    &   \multicolumn{4}{c|}{PSNR $\uparrow$} & \multicolumn{4}{c|}{SSIM $\uparrow$} & \multicolumn{4}{c}{LPIPS $\downarrow$}  \\
    & Near & Medium & Far & Avg. & Near & Medium & Far & Avg. & Near & Medium & Far & Avg. \\ \hline
    3DGS~\cite{kerbl3Dgaussians}&\cellcolor{orange}27.20 &32.57 &37.12 &32.30 &\cellcolor{tablered}0.906 &0.970 &0.995 &\cellcolor{orange}0.957 &\cellcolor{orange}0.150 &0.031 &0.005 &\cellcolor{orange}0.062
\\
Mipmap-GS~\cite{mipmapgs}&26.59 &31.61 &34.66 &30.95 &0.898 &0.967 &0.992 &0.952 &0.157 &0.034 &0.005 &0.066
\\
Multiscale-3DGS\cite{multiscale-3dgs} &26.31 &31.54 &34.66 &30.84 &0.898 &0.967 &0.992 &0.952 &0.158 &0.034 &0.005 &0.066
\\
Mip-Splatting~\cite{mip-splatting} &26.87 &\cellcolor{yellow}33.08 &\cellcolor{orange}40.76 &\cellcolor{yellow}33.57 &\cellcolor{yellow}0.902 &\cellcolor{tablered}0.973 &\cellcolor{tablered}0.997 &\cellcolor{orange}0.957 &\cellcolor{yellow}0.154 &\cellcolor{tablered}0.029 &\cellcolor{tablered}0.003 &\cellcolor{orange}0.062
\\
Analytic-Splatting~\cite{analytic-splatting} &\cellcolor{yellow}27.06 &\cellcolor{orange}33.27 &\cellcolor{yellow}40.41 &\cellcolor{orange}33.58 &0.901 &\cellcolor{orange}0.972 &\cellcolor{tablered}0.997 &\cellcolor{orange}0.957 &\cellcolor{yellow}0.154 &\cellcolor{orange}0.030 &\cellcolor{tablered}0.003 &\cellcolor{orange}0.062
\\
\hline
LOD-GS (ours) &\cellcolor{tablered}27.30 &\cellcolor{tablered}33.40 &\cellcolor{tablered}41.27 &\cellcolor{tablered}33.99 &\cellcolor{orange}0.905 &\cellcolor{tablered}0.973 &\cellcolor{tablered}0.997 &\cellcolor{tablered}0.958 &\cellcolor{tablered}0.148 &\cellcolor{orange}0.030 &\cellcolor{tablered}0.003 &\cellcolor{tablered}0.060

% w/o EWA filter &27.24 &32.77 &38.08 &32.70 &0.907 &0.971 &0.996 &0.958 &0.147 &0.031 &0.004 &0.061
% \\
% w/o LOD filter &27.28 &33.30 &40.45 &33.68 &0.905 &0.972 &0.997 &0.958 &0.151 &0.030 &0.003 &0.061
    \end{tabular}
    }
    \vspace{-0.5em}
    \caption{
    \textbf{Multi-level Training and Multi-level Testing on our extended Blender dataset.} Our LOD-GS achieves state-of-the-art performance in multi-level scene representation compared to vanilla 3DGS and existing anti-aliasing Gaussian splatting methods.
    }
    \label{tab:avg_multilevel_results}
    \vspace{-1em}
\end{table*}

%-----------------------------------------------------------------------------
%\vspace{-2mm}
\section{Experimental Evaluation}
\label{sec:experiment}
%\newcommand{\tci}[1]{\tiny{~\cite{#1}}}
%-----------------------------------------------------------------------------

\subsection{Implementation}
%general intro
% Our LOD-GS follows the framework of the vanilla 3DGS.
The LOD filtering is implemented using PyTorch and is designed to be plug-and-play, allowing for easy integration into existing 3DGS-based methods.
It is implemented using GMM and we set the number of basis functions to 20 in the following experiment.
We evaluate our method on both synthetic and real-world datasets, including the Multi-scale Synthetic Dataset from Mip-NeRF~\cite{mip-nerf} and the Mip-NeRF~360~\cite{mipnerf360} dataset.
We render a multi-level synthetic dataset for more comprehensive evaluation as illustrated in Section~\ref{subsec:multi-level}.
We also evaluate the generalization ability of our method in Section~\ref{subsec:stmt} and examine the impact of the number of basis functions used in the LOD filter in Section~\ref{subsec:basis}.
In our experiment, Mip-Splatting uses the densification scheme from GOF~\cite{GOF}, while all other methods, including LOD-GS, utilize the original densification scheme of 3DGS.
All experiments are conducted on a single RTX3090 GPU.

\subsection{Evaluation on the Multi-scale Blender Dataset}
\label{subsec:ms-blender}

The Blender dataset introduced in the original NeRF~\cite{mildenhall2020nerf} is a synthetic dataset where all training and testing images observe the scene content from a roughly constant distance with the same focal length, which differs significantly from real-world captures.
MipNeRF~\cite{barron2021mip} introduces a multi-scale Blender dataset designed to enhance the evaluation of reconstruction accuracy and anti-aliasing in multi-resolution scenes. This dataset is generated by downscaling the original dataset by factors of 2, 4, and 8, and then combining these variations.
We evaluate our LOD-GS on this dataset with several competitive methods, including NeRF-based methods (NeRF~\cite{mildenhall2020nerf}, MipNeRF~\cite{barron2021mip}, and Tri-MipRF~\cite{trimip-nerf}) and 3DGS-based methods (Mipmap-GS~\cite{mipmapgs}, SA-GS~\cite{sa-gs}, Multiscale-3DGS~\cite{multiscale-3dgs}, Mip-Splatting~\cite{mip-splatting}, and Analytic-Splatting~\cite{analytic-splatting}). Following previous works, we report three metrics: PSNR, SSIM~\cite{wang2004image}, and VGG LPIPS~\cite{zhang2018unreasonable} across four resolutions and average results, as shown in Table~\ref{tab:avg_multiblender_results}.
Mip-NeRF and Tri-MipRF exhibit stronger anti-aliasing capabilities compared to vanilla NeRF. For 3DGS-based methods, Mip-Splatting and Analytic-Splatting significantly enhance the anti-aliasing ability of 3DGS. While Multiscale-3DGS can achieve better performance at the original resolution, it degenerates at lower resolutions. 
It is important to emphasize that both Mip-Splatting and Analytic-Splatting require sampling more cases from high resolutions to maintain their performance at the original resolution. 
In contrast, by utilizing LOD-sensitive filter control, our method achieves rather good performance without relying on any dataset resampling trick.
As shown in Table~\ref{tab:avg_multiblender_results}, our LOD-GS achieves state-of-the-art anti-aliasing performance while maintaining competitive rendering quality at the original resolution.

\begin{figure}[t] 
	\centering
    \includegraphics[width=\linewidth]{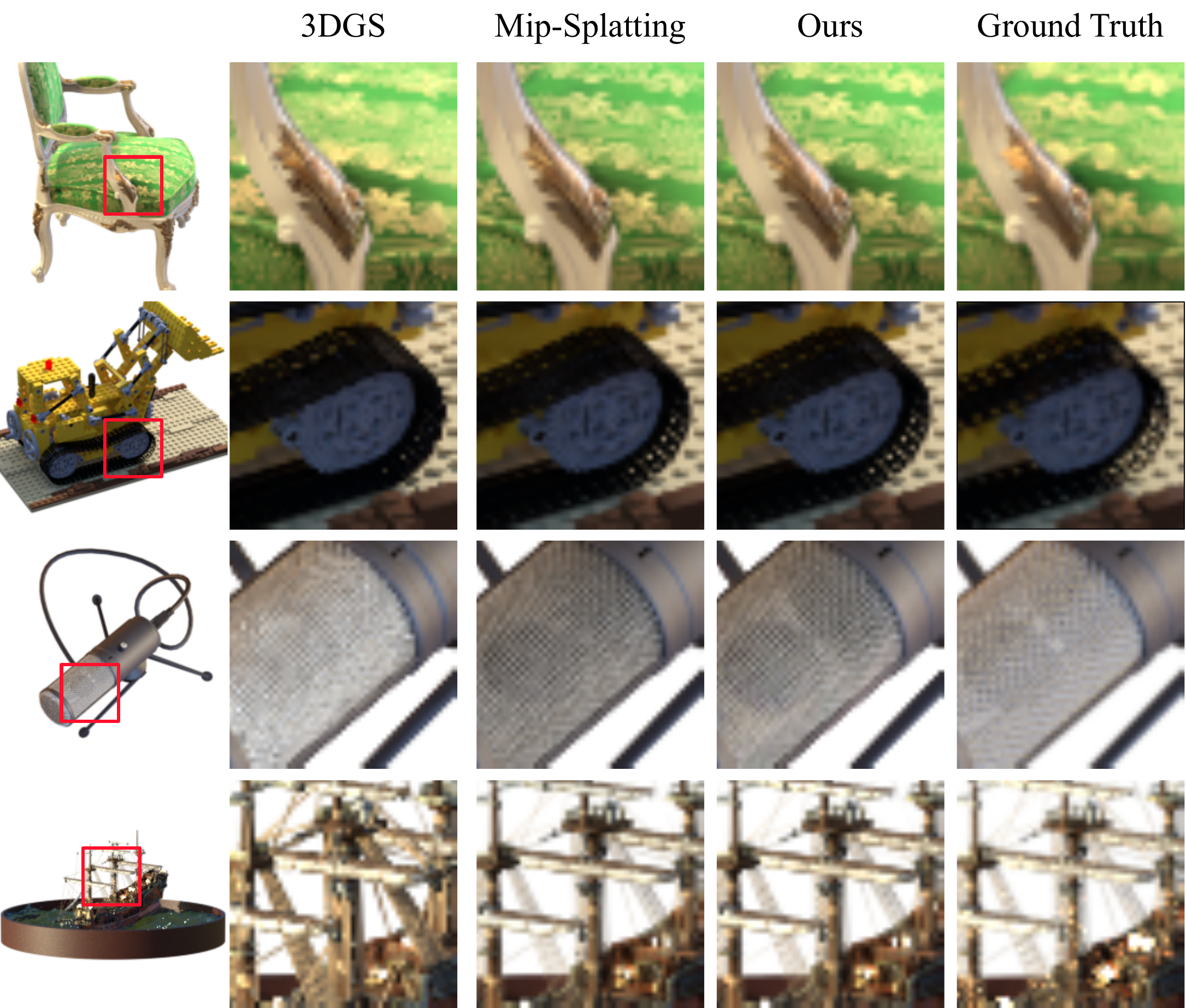}
	% \vspace{1mm}
	\caption{
    \textbf{Qualitative comparison on Anti-aliasing.}
    We compare the results of our method with the aliased rendering results of 3DGS and the anti-aliased results of Mip-Splatting to illustrate the anti-aliasing capabilities of our proposed method.
    The positions of the details on the original objects are marked with red boxes.
    }
	\vspace{-0.5em}
	\label{fig:filter-cmp}
\end{figure}

\subsection{Evaluation on the Multi-level Blender Dataset}
\label{subsec:multi-level}
Both decreasing the focal length and increasing the camera distance can change the sampling rate, thereby reducing the resolution of the photographed object.
Mip-NeRF reduces image resolution using image processing techniques, attributing this change to alterations in focal length. 
However, in 3D applications, the most common approach to change the sampling rate is by adjusting the camera distance, rather than the focal length. 
There is a difference between adjusting the camera distance and the focal length. 
To make a more comprehensive evaluation, we re-render the synthetic dataset used in NeRF~\cite{mildenhall2020nerf} from three different camera distances. 
We call this extended dataset the ``Multi-level Blender Dataset" and assign Level 1 to the camera with the nearest camera distance to the object, Level 2 to the middle camera distance, and Level 3 to the farthest distance.
The results of the comparative experiments on this dataset are presented in Table~\ref{tab:avg_multilevel_results}.
As shown, 3DGS performs well when the object is close to the camera but degrades in performance when the camera is far from the object. Conversely, Mip-Splatting and Analytic-Splatting exhibit better performance at greater distances but struggle at near distances. Only LOD-GS achieves good performance at both near and far distances.
It is crucial to note that when the camera is positioned at a far distance, a significant portion of the image appears white. This can lead to smaller differences in SSIM and LPIPS between various methods. In this context, PSNR is a more appropriate metric, as it is calculated based on pixel differences.

\myparagraph{Qualitative results.}
We further conducted a qualitative comparison of our LOD-GS with vanilla 3DGS and existing GS-based methods for the reconstruction of details. 
As shown in Figure~\ref{fig:detail-cmp}, both 3DGS and Multiscale-3DGS fail to capture very fine details, resulting in noticeable discrepancies in appearance compared to the ground-truth images. Although Mip-Splatting and Analytic-Splatting retain some texture detail, they still lose additional intricacies and produce overly blurred rendering results in certain areas. 
In contrast, our method captures more details and achieves higher rendering quality.
To be more precise, as illustrated in Figure~\ref{fig:detail-cmp}, our method excels at reconstructing the small bump on the back of the chair, the intricate texture on the surface of the microphone, and the rope net of the ship.
In contrast, other methods either fail to capture these fine details or produce overly blurred results.

In addition to capturing fine details, we also evaluate the anti-aliasing capabilities of our method. We render the object from a considerable distance to assess this ability. The qualitative results are illustrated in Figure~\ref{fig:filter-cmp}. 
As shown, 3DGS tends to generate dilated boundaries and aliased results in areas with high-frequency texture. 
In contrast, both Mip-Splatting and our LOD-GS effectively handle aliasing when the camera is far from the scene, i.e., at a low sampling rate.
Compared to Mip-Splatting, our method can reconstruct more details, such as the inner structure of the microphone and the rope net of the ship, as shown in the third row and the fourth row of the Figure~\ref{fig:filter-cmp}.

% \renewcommand{\arraystretch}{1.1}
% \begin{table}[t]
%     \renewcommand{\tabcolsep}{3pt}
%     \centering
%     \resizebox{0.95\linewidth}{!}{
%     \begin{tabular}{@{}l@{\,\,}|cccc|cccc|}
%     &   \multicolumn{4}{c|}{w/o EWA Filter} & \multicolumn{4}{c|}{w/o LOD Control}  \\
%     & Level 1 & Level 2 & Level 3 & Avg. & Level 1 & Level 2 & Level 3 & Avg. \\ \hline
%     \input{tab/ablation_level.tex}
%     \end{tabular}
%     }
%     \vspace{0.1in}
%     \caption{
%     Ablation study on our extended Blender dataset which contains three different levels (Level 1: near, Level 2: middle, Level 3: far). We display the performance of model after removing the EWA filter and our LOD control module respectively.
%     }
%     \label{tab:ablation_level}

% \end{table}

% \renewcommand{\arraystretch}{1.1}
% \begin{table}[t]
%     \renewcommand{\tabcolsep}{3pt}
%     \centering
%     \resizebox{\linewidth}{!}{
%     \begin{tabular}{@{}l@{\,\,}|cccc|cccc|}
%     &   \multicolumn{4}{c|}{w/o EWA Filter} & \multicolumn{4}{c|}{w/o LOD Control}  \\
%     & Full Res. & $\nicefrac{1}{2}$ Res. & $\nicefrac{1}{4}$ Res. & $\nicefrac{1}{8}$ Res. & Full Res. & $\nicefrac{1}{2}$ Res. & $\nicefrac{1}{4}$ Res. & $\nicefrac{1}{8}$ Res.\\ \hline
%     \input{tab/ablation_scale.tex}
%     \end{tabular}
%     }
%     \vspace{0.1in}
%     \caption{
%     Ablation study on multi-scale Blender dataset. We display the performance of model after removing the EWA filter and our LOD control module respectively.
%     }
%     \label{tab:ablation_scale}
% \end{table}

\subsection{Evaluation on the Mip-NeRF 360 Dataset}
\begin{figure}[!t] 
	\centering
    \includegraphics[width=\linewidth]{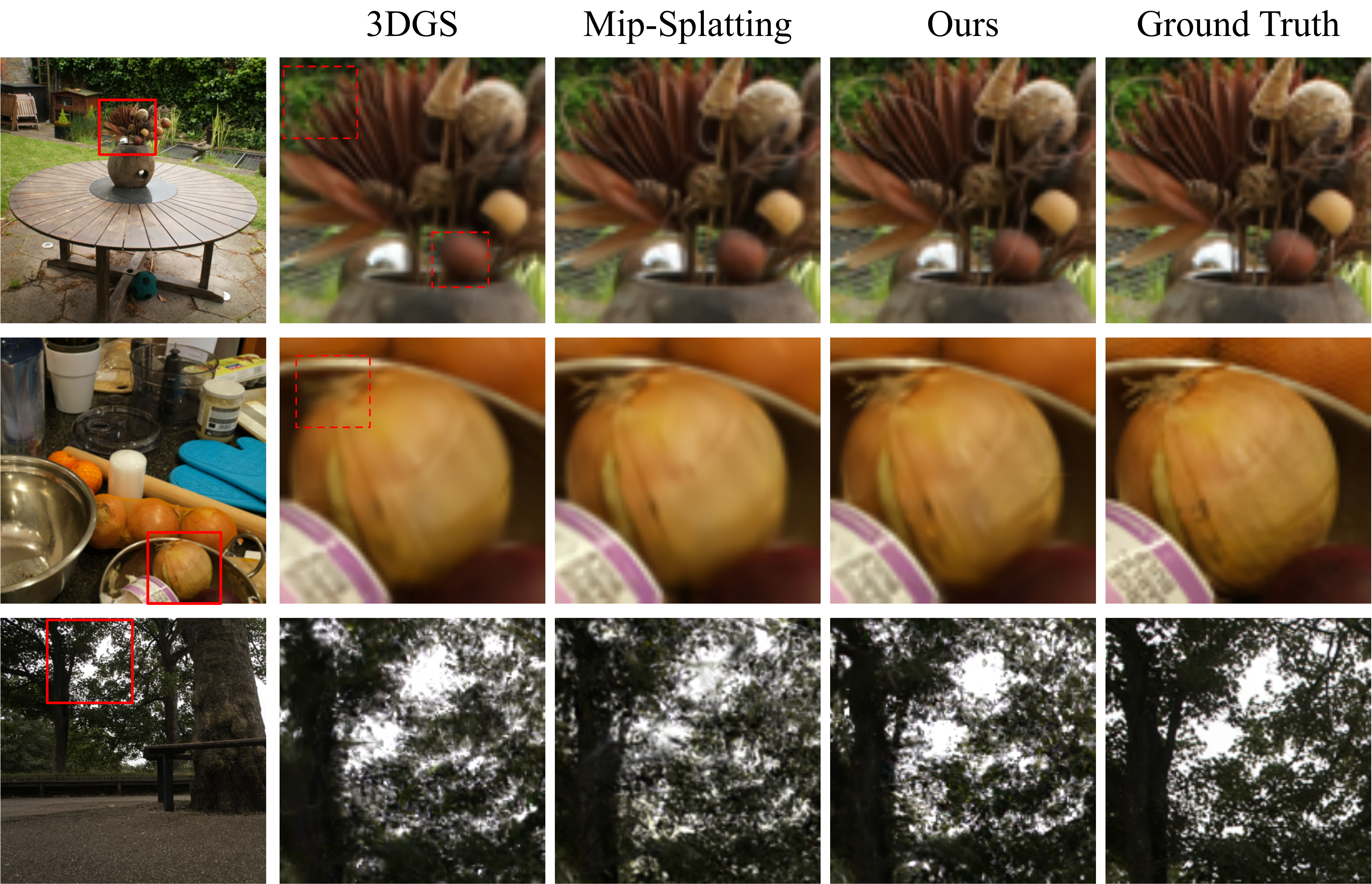}
	\caption{
    \textbf{Qualitative comparison on the Mip-NeRF 360 dataset~\cite{mipnerf360}.}
    All methods are trained and tested at three downsampled resolutions ($\nicefrac{1}{8}$, $\nicefrac{1}{16}$, and $\nicefrac{1}{32}$).
    We recommend that readers scale up this image and compare the region marked with red boxes across different methods.
    }
    % \vspace{-1em}
	\label{fig:369cmp}
\end{figure}

To test the performance of our method on real-world data, we evaluate our method and perform comparisons on the Mip-NeRF 360 Dataset~\cite{mipnerf360}. 
Due to the high resolution of images in the 360 Dataset and the need to perform multiscale training, we conduct multiscale training and testing on three downsampled scales (\(\nicefrac{1}{8}\), \(\nicefrac{1}{16}\), and \(\nicefrac{1}{32}\)) to reduce the VRAM consumption. 
The qualitative and quantitative experimental results are shown in Figure~\ref{fig:369cmp} and Table~\ref{tab:multi_360}. 
As shown in Table~\ref{tab:multi_360}, our method achieves SOTA performance across nearly all metrics at different resolutions.

\renewcommand{\arraystretch}{1.1}
\begin{table*}[!t]
    \renewcommand{\tabcolsep}{1pt}
    \centering
    \resizebox{0.95\linewidth}{!}{
    \begin{tabular}{@{}l@{\,\,}|cccc|cccc|cccc}
    &   \multicolumn{4}{c|}{PSNR $\uparrow$} & \multicolumn{4}{c|}{SSIM $\uparrow$} & \multicolumn{4}{c}{LPIPS $\downarrow$}  \\
    &  $\nicefrac{1}{8}$ Res. & $\nicefrac{1}{16}$ Res. & $\nicefrac{1}{32}$ Res. & Avg. &  $\nicefrac{1}{8}$ Res. & $\nicefrac{1}{16}$ Res. & $\nicefrac{1}{32}$ Res. & Avg. &  $\nicefrac{1}{8}$ Res. & $\nicefrac{1}{16}$ Res. & $\nicefrac{1}{32}$ Res. & Avg. \\ \hline
    3DGS~\cite{kerbl3Dgaussians} &27.99 &29.96 &29.62 &29.19 &0.850 &0.919 &0.934 &0.901 &0.158 &0.074 &0.057 & 0.096
\\
Mipmap-GS~\cite{mipmapgs} &27.28 &29.23 &28.50 &28.33 &0.836 &0.909 &0.919 &0.888 &0.173 &0.084 &0.066 &0.108
\\
Multiscale-3DGS\cite{multiscale-3dgs} & 27.30 &29.25 &28.50 &28.35 &0.836 &0.910 &0.920 &0.888 &0.173 &0.084 &0.066 &0.108
\\
Mip-Splatting~\cite{mip-splatting} &\cellcolor{yellow}28.35 &\cellcolor{yellow}30.16 &\cellcolor{yellow}31.06 &\cellcolor{yellow}29.86 &\cellcolor{yellow}0.865 &\cellcolor{yellow}0.924 &\cellcolor{yellow}0.949 &\cellcolor{yellow}0.913 &\cellcolor{orange}0.132 &\cellcolor{tablered}0.064 &\cellcolor{orange}0.043 &\cellcolor{yellow}0.080
\\
Analytic-Splatting~\cite{analytic-splatting} &\cellcolor{orange}28.84 &\cellcolor{orange}30.63 &\cellcolor{orange}31.63 &\cellcolor{orange}30.37 &\cellcolor{orange}0.868 &\cellcolor{orange}0.926 &\cellcolor{orange}0.953 &\cellcolor{orange}0.915 &\cellcolor{tablered}0.128 &\cellcolor{yellow}0.066 &\cellcolor{orange}0.043 &\cellcolor{orange}0.079
\\

\hline
LOD-GS (ours) &\cellcolor{tablered}28.99 &\cellcolor{tablered}30.93 &\cellcolor{tablered}32.48 &\cellcolor{tablered}30.80 &\cellcolor{tablered}0.870 &\cellcolor{tablered}0.929 &\cellcolor{tablered}0.958 &\cellcolor{tablered}0.919 &\cellcolor{yellow}0.133 &\cellcolor{orange}0.065 &\cellcolor{tablered}0.038 &\cellcolor{tablered}0.078

% \cellcolor{tablered}
% \cellcolor{yellow}
% \cellcolor{orange}
    \end{tabular}
    }
    \vspace{-0.5em}
    \caption{
    \textbf{Multi-scale training and Multi-scale testing on the Mip-NeRF 360 dataset~\cite{mipnerf360}.} 
    All methods are trained and tested at three downsampled resolutions ($\nicefrac{1}{8}$, $\nicefrac{1}{16}$, and $\nicefrac{1}{32}$). Our model achieves state-of-the-art performance across most metrics in comparison.
    }
    \label{tab:multi_360}
    \vspace{-0.5em}
\end{table*}

\begin{figure*}[!t] 
	\centering
    \includegraphics[width=\linewidth]{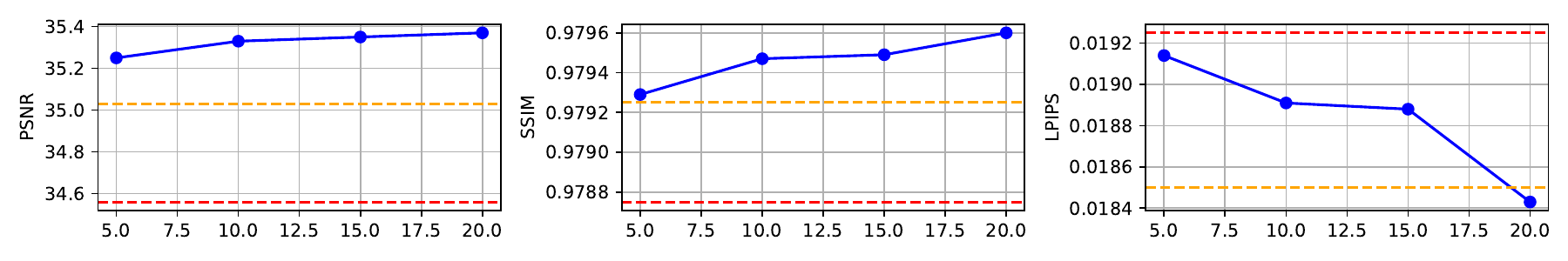}
    \vspace{-2em}
	\caption{
    \textbf{Performance of Models Using Different Numbers of Basis Functions:} We report \textcolor{blue}{our model's} changes in PSNR, SSIM, and LPIPS by using 5, 10, 15, and 20 basis functions. Metrics for \textcolor{red}{Mip-Splatting} and \textcolor{dark_orange}{Analytic-Splatting} are visualized with dashed lines.
    }
    \vspace{-1.5em}
	\label{fig:basis}
\end{figure*} 

\begin{table}[t]
    \centering
    \begin{minipage}{0.5\columnwidth}
    \centering
    \footnotesize
    \caption{Ablation on Multiscale Blender dataset.}
    \footnotesize
    \vspace{-1em}
    \setlength{\tabcolsep}{2.0pt}
    \begin{tabular*}{1.0\columnwidth}{cccc}
        \toprule
         & PSNR$\uparrow$ & SSIM$\uparrow$& LPIPS$\downarrow$\\
        \midrule
        w/o LOD& 34.39& 0.977& 0.022\\ 
        w/o EWA& 34.61& 0.978& 0.020 \\
        full model& 35.37& 0.980& 0.018\\
        \bottomrule
    \end{tabular*}
    \label{tab:ablation}
    \end{minipage}%
    \hspace{0.05\columnwidth}
    \begin{minipage}{0.4\columnwidth}
    \centering
    \setlength{\tabcolsep}{0.1pt}
    \footnotesize
    \caption{STMT on Multi-level Blender dataset.}
    \footnotesize
    \vspace{-1em}
    \begin{tabular}{>{\centering\arraybackslash}p{30pt} >{\centering\arraybackslash}p{25pt} >{\centering\arraybackslash}p{25pt} >{\centering\arraybackslash}p{25pt}}
        \toprule
         PSNR $\uparrow$& Zoom In  & Trained Scale& Zoom Out\\
        \midrule
        3DGS& 23.38 & 35.73& 37.85\\ 
        Mip& 23.50& 35.30& 39.54\\ 
        Analytic\footnotesize& 23.40& 34.66& 40.09\\
        LOD-GS&\textbf{24.27} &\textbf{35.95} &\textbf{41.42} \\
        \bottomrule
    \end{tabular}
    \label{tab:stmt}
    \end{minipage}
    \vspace{-1.4em}
\end{table}

\myparagraph{Qualitative Results}
To provide a more intuitive comparison of real-world scenes, we analyze the rendering results of different methods across various scenes, as shown in Figure~\ref{fig:369cmp}. Compared to 3DGS and Mip-Splatting, the most notable distinction of our method is its ability to reconstruct highly detailed geometry and texture following multiscale training. For instance, in Figure~\ref{fig:369cmp}, our method effectively reconstructs the microstructure of the plant in the \emph{garden} scene, the texture of the onion in the \emph{counter} scene, and the complex distribution of leaves in the \emph{treehill} scene. All these results demonstrate that our method achieves high-quality rendering after multiscale training.

\subsection{Ablation Study}
We remove the LOD filtering and EWA filter, respectively, to validate the effectiveness of these two components.
The ablation study is conducted on the Multiscale Blender Dataset and the experiment result is presented in Table \ref{tab:ablation}.
Only using the LOD filtering can achieve rather good performance while utilizing EWA filter can further improve it.
Because the EWA filter can sense the change in sampling rate to some degree, utilizing it can decrease the learning burden of LOD filtering, thus achieving better performance using the same number of training iterations.

\subsection{Generalization on STMT}
\label{subsec:stmt}
Our method is primarily designed for Multiscale Training and Multiscale Testing (MTMT).
To test the generalization of our method, we also perform the Single-scale Training and Multiscale Testing (STMT) on the Multi-level Blender dataset.
As shown in Table \ref{tab:stmt}, our method achieves the best performance across all scales, even if we only train it using one scale.
The experimental result demonstrates that our method can generalize to STMT well.

\subsection{Analysis of the Number of Basis Functions}
\label{subsec:basis}
We investigate the influence of the number of basis functions on the model's performance and present the experimental results in Figure~\ref{fig:basis}. As shown, using only 5 basis functions allows our method to surpass both Mip-Splatting and Analytic-Splatting in terms of PSNR and SSIM. With an increase in the number of basis functions, our model exhibits improved performance across PSNR, SSIM, and LPIPS; however, this also results in higher computational costs. This observation provides a guideline for selecting the number of basis functions: opting for fewer functions for efficiency or more functions for enhanced performance.

%%\vspace{-1mm}
%-----------------------------------------------------------------------------
\section{Conclusion}
\label{sec:conclusion}
\vspace{-0.5em}
In this paper, we propose a Level-of-Detail-Sensitive Gaussian Splatting method, LOD-GS, to sense and learn the filtering degree changes caused by variations in the camera sampling rate.
Compared to existing methods, our approach achieves state-of-the-art performance on anti-aliasing tasks while preserving very fine details. 
This work also re-renders the Synthetic NeRF dataset from different camera distances to facilitate the comprehension evaluation of the anti-aliasing task.
Experimental results also demonstrate that LOD-GS generalize well to STMT and can achieve rather good performance with minor additional parameters.
LOD-GS is designed for easy reproduction and integration.
It is expected to further improve the reconstruction quality of neural radiance fields, especially when the collected images are captured at varying sampling rates.

{
    \small
    \clearpage
    \bibliographystyle{ieee_3dv}
    \bibliography{egbib}
}

\clearpage
\setcounter{page}{1}
\newpage

\twocolumn[
\centering
\Large
\textbf{LOD-GS: Level of Detail-Sensitive 3D Gaussian Splatting for \\Detail Conserved Anti-Aliasing}\\
\vspace{1em}
Supplementary Material\\
\vspace{2.0em}
]

% \twocolumn[
% \centering
% \Large
% Supplementary Material\\
% \vspace{0.5em}
% \vspace{1.0em}
% ]

In this supplementary material, we provide a comprehensive ablation study and its analysis in Section~\ref{sec:ablation}. We also report the complete results of comparison experiments conducted on the Multi-scale Blender Dataset~\cite{mip-nerf}, our newly rendered Multi-level Blender Dataset, and the Mip-NeRF 360 Dataset~\cite{mipnerf360} in Section~\ref{sec:add_res}.

\section{Ablation}
\label{sec:ablation}
The capability of our method to achieve aliasing-free and detail-conserved rendering relies on two components: the level-of-detail sensitive filter (denoted as LOD filter hereafter) and the EWA filter. 
To assess the contribution of these modules to overall performance of our method, we separately remove the LOD and EWA filters and conduct model training on synthetic and real-world dataset. Detailed analyses of these two components are presented in Section~\ref{sec:lod} and Section~\ref{sec:ewa}.

\subsection{Effectiveness of the LOD filter}
\label{sec:lod}
\begin{figure*}[h] 
	\centering
    \includegraphics[width=0.93\linewidth]{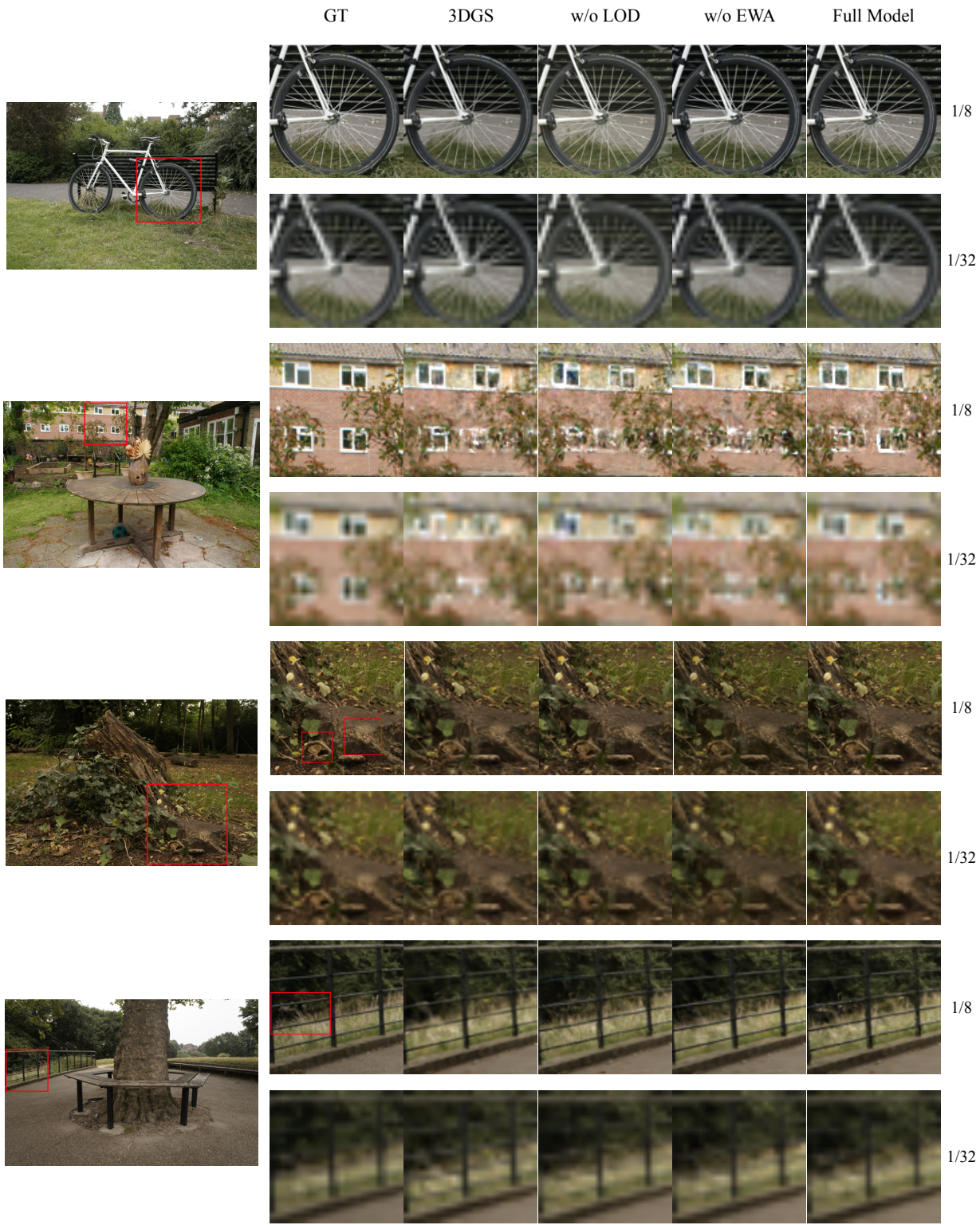}
	\caption{
    \textbf{Qualitative Results of the Ablation Study on the Mip-NeRF 360 Dataset~\cite{mipnerf360}.} All methods are trained and tested on multiscale inputs ($\nicefrac{1}{8}$, $\nicefrac{1}{16}$, $\nicefrac{1}{32}$). We present rendering results from different methods at $\nicefrac{1}{8}$ and $\nicefrac{1}{32}$ scales to evaluate detail conservation and anti-aliasing performance.
}
	\label{fig:supp_mipnerf}
\end{figure*}

The LOD filter is mainly responsible for disentangling inputs with different sampling rates.
It helps alleviate the ambiguity caused by the multi-scale input, which is the key to learn a radiance field with fine details and anti-aliasing ability at the same time.
As shown in Figure~\ref{fig:supp_blender}, scenes trained without the LOD filter fail to capture the fine details of the objects. For instance, the inner structure of the microphone is unclear when photographed from a near distance. 
Other details, such as the metal accents and rope net of the ship,  as well as the bumps on the chair, cannot be well reconstructed without the LOD filter.
Moreover, as illustrated in Figure~\ref{fig:supp_mipnerf}, using only the LOD filter can mitigate aliasing to some degree compared to the vanilla 3DGS~\cite{kerbl3Dgaussians}. Experimental results in Table~\ref{tab:ablation_blender} and Table~\ref{tab:ablation_mip360} demonstrate that removing the LOD filter leads to a more significant performance drop compared to removing the EWA filter.
All these results illustrate that the LOD filter in our method is crucial for anti-aliasing and the conservation of fine details.
\subsection{Effectiveness of the EWA filter}
\label{sec:ewa}
\begin{figure*}[h] 
	\centering
    \includegraphics[width=0.9\linewidth]{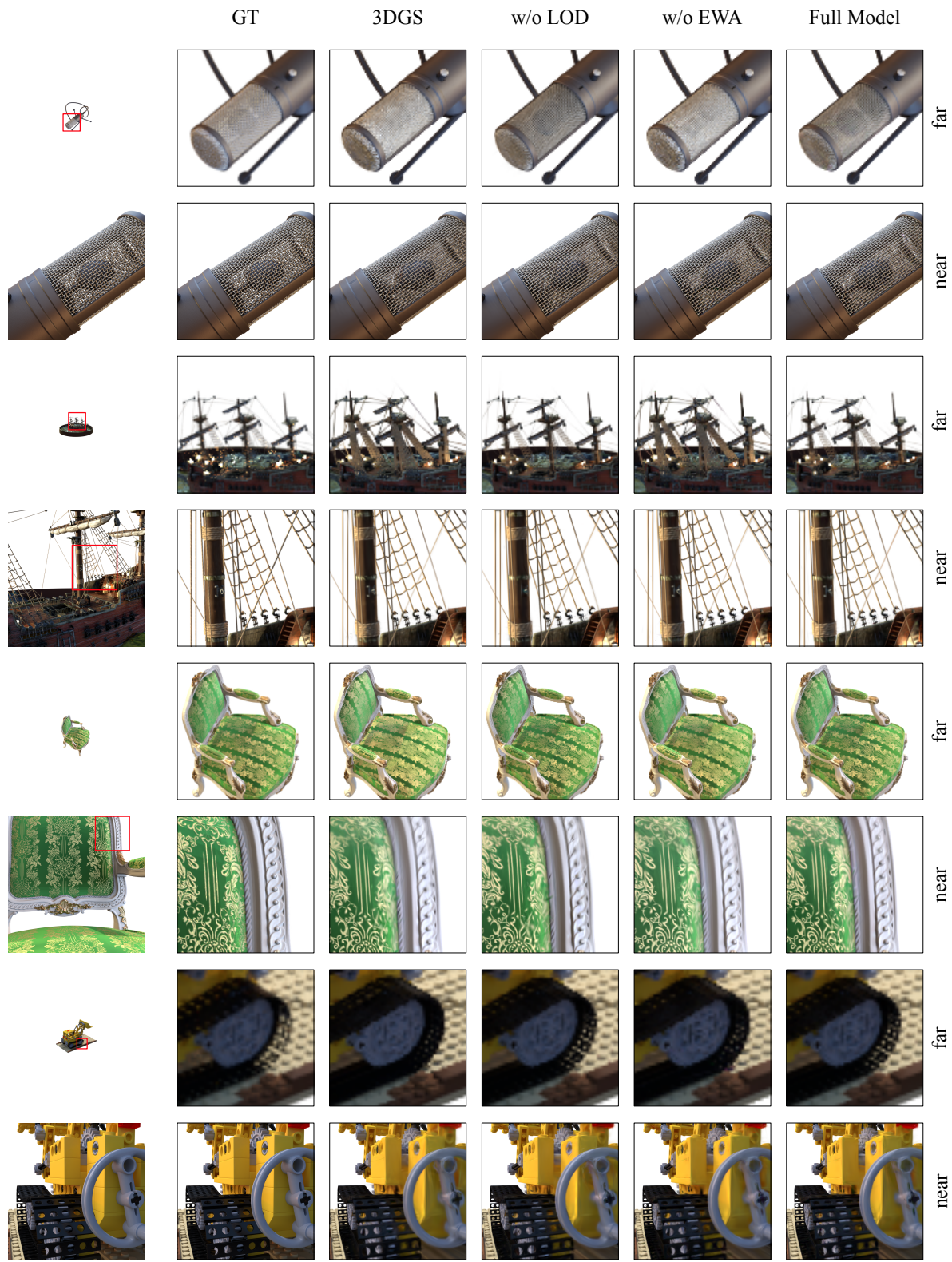}
	\caption{
    \textbf{Qualitative Results of the Ablation Study on the Blender Dataset~\cite{mildenhall2020nerf, mip-nerf}.} We render the objects from both far and near distances to test the anti-aliasing and detail conservation abilities.
    }
	\label{fig:supp_blender}
\end{figure*}

The EWA filter is utilized to enhance the model's anti-aliasing capability. Although our LOD filter can address the aliasing problem to some extent, integrating the EWA filter into our method can provide improved anti-aliasing performance.
As shown in Figure~\ref{fig:supp_blender}, the results rendered from a far distance using the method that only utilizes the LOD filter face issues of aliasing and dilation artifacts. More specifically, the result for \textit{mic} is aliased, while the results for \textit{ship} and \textit{lego} exhibit dilation artifacts.
The quantitative results in Table~\ref{tab:ablation_blender} and Table~\ref{tab:ablation_mip360} illustrate that removing the EWA filter also leads to a performance drop. 
These observations and results demonstrate that the EWA filter contributes to the anti-aliasing capability of our method.

\section{Additional Results}
\label{sec:add_res}
In this section, we report the detail of our comparison experiment on each scene across different sampling rates.
We analyze the experimental results on the Blender Dataset in Section~\ref{sec:blender} and the Mip-NeRF 360 Dataset in Section~\ref{sec:mip360}.

\subsection{Blender Dataset}
\label{sec:blender}
We perform comparison experiments on the Multi-Scale Blender Dataset~\cite{mip-nerf} and our newly rendered Multi-Level Dataset. The Multi-Level Blender dataset does not contain the \textit{drum} scene because of the wrong specular reflection effect in the \textit{drums} blender project.
The experimental results on the Multi-Scale Blender dataset are shown in Table~\ref{tab:multiscale_blender_supp}. Compared to existing powerful methods for radiance field anti-aliasing, such as Mip-NeRF~\cite{mip-nerf}, Mip-Splatting~\cite{mip-splatting}, and Analytic-Splatting~\cite{analytic-splatting}, our method achieves state-of-the-art performance in most scenes.
We collected a new dataset called the Multi-Level Blender dataset, which simulates changes in sampling rate using varying camera distances. The experimental results on this new dataset are reported in Table~\ref{tab:multilevel_blender_supp}. We compare our method with Gaussian Splatting-based methods designed to handle multiscale inputs, including Mipmap-GS~\cite{mipmapgs}, MSGS~\cite{multiscale-3dgs}, Mip-Splatting~\cite{mip-splatting}, and Analytic-Splatting~\cite{analytic-splatting}. 
As shown in Table~\ref{tab:multilevel_blender_supp}, our method achieves the best performance in most scenes. 
These experimental results demonstrate the state-of-the-art performance of our method in handling inputs with varying sampling rates.

\subsection{Mip-NeRF 360 Dataset}
\label{sec:mip360}
We down-sample the full resolution images in Mip-NeRF 360 dataset in three scales (\(\nicefrac{1}{8}\), \(\nicefrac{1}{16}\) and \(\nicefrac{1}{32}\)).
All methods are trained and tested on these three scales.
The experimental results are displayed in the Table~\ref{tab:multiscale_360_supp}.
As shown, our proposed LOD-GS achieves the best or nearly the best performance across all scenes.
This illustrates that our method is also applicable to wild and unbounded real-world data.

\begin{table*}[h]
    \centering
    \small
    \begin{tabular}{l|cccccccc|c}
        & \multicolumn{8}{c}{\textbf{PSNR}} \\
 & \scenename{chair}& \scenename{drums} & \scenename{ficus}  & \scenename{hotdog}  & \scenename{lego}  & \scenename{materials}  & \scenename{mic}  & \scenename{ship} & \scenename{Average} 
 \\ 
 \hline 
Full Model Full Res. &34.99 & 25.95 & 35.06 & 37.28 & 34.76 & 29.66 & 34.90 & 30.61 & \textbf{32.90} \\
Full Model $\nicefrac{1}{2}$ &38.49 & 27.16 & 35.94 & 39.44 & 36.43 & 31.21 & 37.78 & 32.59 & \textbf{34.88} \\
Full Model $\nicefrac{1}{4}$&40.39 & 28.65 & 36.38 & 40.99 & 37.23 & 33.22 & 40.23 & 34.31 & \textbf{36.43}\\
Full Model $\nicefrac{1}{8}$&41.37 & 30.28 & 36.26 & 41.77 & 36.84 & 35.10 & 40.84 & 35.70 & \textbf{37.27}\\
\hline
w/o LOD Full Res.&33.23 & 25.73 & 34.08 & 36.54 & 33.29 & 29.04 & 33.44 & 29.70 & 31.88 \\
w/o LOD $\nicefrac{1}{2}$&37.04 & 27.02 & 35.20 & 38.90 & 35.35 & 30.79 & 36.46 & 31.92 & 34.09 \\
w/o LOD $\nicefrac{1}{4}$&39.43 & 28.51 & 35.95 & 40.56 & 36.45 & 33.05 & 39.07 & 33.89 & 35.86 \\
w/o LOD $\nicefrac{1}{8}$&38.68 & 29.78 & 35.53 & 40.24 & 34.77 & 33.79 & 38.38 & 34.69 & 35.73 \\
\hline
w/o EWA Full Res.&34.59 & 25.88 & 34.51 & 37.15 & 34.44 & 29.62 & 34.17 & 30.66 & 32.63 \\
w/o EWA $\nicefrac{1}{2}$&37.77 & 26.98 & 34.49 & 39.29 & 35.96 & 31.14 & 36.95 & 32.58 & 34.39 \\
w/o EWA $\nicefrac{1}{4}$&39.31 & 28.27 & 33.89 & 40.81 & 36.45 & 32.88 & 39.55 & 34.11 & 35.66\\
w/o EWA $\nicefrac{1}{8}$&38.93 & 29.24 & 33.67 & 40.79 & 35.21 & 33.96 & 39.46 & 34.80 & 35.76\\
% -------------------------------------------------------------------
 \multicolumn{8}{c}{} \\
 & \multicolumn{8}{c}{\textbf{SSIM}} \\
 & \scenename{chair}  & \scenename{drums}& \scenename{ficus}  & \scenename{hotdog}  & \scenename{lego}  & \scenename{materials}  & \scenename{mic}  & \scenename{ship} & \scenename{Average} 
 \\
 \hline
Full Model Full Res. &0.983 & 0.950 & 0.986 & 0.983 & 0.979 & 0.958 & 0.990 & 0.901 & \textbf{0.966}\\
Full Model $\nicefrac{1}{2}$ &0.992 & 0.959 & 0.992 & 0.990 & 0.988 & 0.974 & 0.993 & 0.927 & \textbf{0.977}\\
Full Model $\nicefrac{1}{4}$&0.995 & 0.968 & 0.994 & 0.993 & 0.992 & 0.987 & 0.995 & 0.952 & \textbf{0.985}\\
Full Model $\nicefrac{1}{8}$&0.997 & 0.978 & 0.994 & 0.996 & 0.994 & 0.994 & 0.997 & 0.969 & \textbf{0.990}\\
\hline
w/o LOD Full Res.&0.973 & 0.946 & 0.984 & 0.981 & 0.973 & 0.954 & 0.986 & 0.894 & 0.961\\
w/o LOD $\nicefrac{1}{2}$&0.989 & 0.958 & 0.991 & 0.989 & 0.986 & 0.973 & 0.991 & 0.924 & 0.975 \\
w/o LOD $\nicefrac{1}{4}$&0.994 & 0.967 & 0.994 & 0.993 & 0.992 & 0.986 & 0.995 & 0.949 & 0.984 \\
w/o LOD $\nicefrac{1}{8}$&0.995 & 0.976 & 0.993 & 0.995 & 0.992 & 0.993 & 0.997 & 0.968 & 0.989 \\
\hline
w/o EWA Full Res.&0.981 & 0.950 & 0.986 & 0.983 & 0.978 & 0.959 & 0.988 & 0.903 & 0.966 \\
w/o EWA $\nicefrac{1}{2}$&0.992 & 0.959 & 0.990 & 0.989 & 0.988 & 0.974 & 0.992 & 0.928 & 0.977 \\
w/o EWA $\nicefrac{1}{4}$&0.995 & 0.966 & 0.990 & 0.993 & 0.992 & 0.986 & 0.995 & 0.950 & 0.983 \\
w/o EWA $\nicefrac{1}{8}$&0.995 & 0.974 & 0.989 & 0.995 & 0.992 & 0.992 & 0.996 & 0.966 & 0.987\\

 % -------------------------------------------------------------------
\multicolumn{8}{c}{} \\
 & \multicolumn{8}{c}{\textbf{LPIPS}} \\
 & \scenename{chair}  & \scenename{drums}& \scenename{ficus}  & \scenename{hotdog}  & \scenename{lego}  & \scenename{materials}  & \scenename{mic}  & \scenename{ship} & \scenename{Average} 
\\ 
\hline
Full Model Full Res. &0.018 & 0.044 & 0.014 & 0.025 & 0.023 & 0.040 & 0.008 & 0.112 & \textbf{0.035}\\
Full Model $\nicefrac{1}{2}$ &0.007 & 0.030 & 0.007 & 0.011 & 0.009 & 0.017 & 0.004 & 0.066 & \textbf{0.019}\\
Full Model $\nicefrac{1}{4}$&0.005 & 0.026 & 0.005 & 0.006 & 0.006 & 0.009 & 0.004 & 0.035 & \textbf{0.012}\\
Full Model $\nicefrac{1}{8}$&0.003 & 0.021 & 0.005 & 0.004 & 0.006 & 0.005 & 0.004 & 0.019 & \textbf{0.008}\\
\hline
w/o LOD Full Res.&0.029 & 0.050 & 0.018 & 0.028 & 0.031 & 0.047 & 0.013 & 0.122 & 0.042\\
w/o LOD $\nicefrac{1}{2}$&0.011 & 0.032 & 0.009 & 0.012 & 0.011 & 0.019 & 0.006 & 0.069 & 0.021 \\
w/o LOD $\nicefrac{1}{4}$&0.006 & 0.027 & 0.006 & 0.006 & 0.007 & 0.010 & 0.004 & 0.037 & 0.013 \\
w/o LOD $\nicefrac{1}{8}$&0.006 & 0.023 & 0.006 & 0.004 & 0.008 & 0.007 & 0.006 & 0.021 & 0.010 \\
\hline
w/o EWA Full Res.&0.022 & 0.045 & 0.014 & 0.027 & 0.025 & 0.040 & 0.010 & 0.119 & 0.038 \\
w/o EWA $\nicefrac{1}{2}$&0.009 & 0.031 & 0.009 & 0.012 & 0.010 & 0.017 & 0.005 & 0.067 & 0.020 \\
w/o EWA $\nicefrac{1}{4}$&0.005 & 0.027 & 0.009 & 0.006 & 0.007 & 0.010 & 0.004 & 0.036 & 0.013 \\
w/o EWA $\nicefrac{1}{8}$&0.004 & 0.024 & 0.009 & 0.004 & 0.008 & 0.007 & 0.005 & 0.021 & 0.010\\
\multicolumn{8}{c}{}
    \end{tabular}
    \caption{
    \textbf{Quantitative Results of the Ablation Study on the Blender Dataset~\cite{mildenhall2020nerf, mip-nerf}.} We present the experimental results of our full model alongside methods that exclude the LOD module or EWA filter across different scenes and resolutions. The results demonstrate that both the LOD and EWA filters significantly contribute to the overall performance of our method.
}
    \label{tab:ablation_blender}
\end{table*}

\begin{table*}[h]
    \centering
    \small
    \begin{tabular}{l|ccccccccc|c}
        & \multicolumn{10}{c}{\textbf{PSNR}} \\
 & \scenename{bicycle} & \scenename{bonsai}  & \scenename{counter}  & \scenename{flowers}  & \scenename{garden}  & \scenename{kitchen}  & \scenename{room} & \scenename{stump} &\scenename{treehill}& \scenename{Average} 
 \\ 
 \hline 
 
Full Model $\nicefrac{1}{8}$ &27.08 & 32.89 & 30.19 & 23.76 & 29.05 & 32.81 & 33.03 & 27.67 & 24.38 & \textbf{28.99} \\
Full Model $\nicefrac{1}{16}$&29.37 & 34.60 & 31.63 & 26.77 & 31.64 & 34.37 & 33.90 & 29.90 & 26.19 & \textbf{30.93} \\
Full Model $\nicefrac{1}{32}$&30.95 & 35.52 & 32.98 & 29.42 & 33.28 & 35.50 & 34.61 & 32.07 & 28.02 & \textbf{32.48} \\
\hline
w/o LOD $\nicefrac{1}{8}$&26.95 & 32.02 & 29.78 & 23.56 & 28.61 & 31.99 & 32.60 & 27.56 & 24.18 & 28.59 \\
w/o LOD $\nicefrac{1}{16}$&29.17 & 33.77 & 31.30 & 26.43 & 31.14 & 33.82 & 33.64 & 29.69 & 26.00 & 30.55 \\
w/o LOD $\nicefrac{1}{32}$&30.23 & 33.86 & 32.13 & 28.77 & 32.30 & 33.90 & 33.72 & 31.41 & 27.57 & 31.54 \\
\hline
w/o EWA $\nicefrac{1}{8}$&27.09 & 32.56 & 29.93 & 23.66 & 28.96 & 32.27 & 32.56 & 27.80 & 24.46 & 28.81 \\
w/o EWA $\nicefrac{1}{16}$&29.29 & 34.25 & 31.26 & 26.54 & 31.51 & 33.76 & 33.42 & 29.74 & 26.15 & 30.66\\
w/o EWA $\nicefrac{1}{32}$&30.41 & 34.44 & 31.87 & 28.50 & 32.75 & 34.08 & 33.76 & 31.16 & 27.42 & 31.60 \\

 \multicolumn{10}{c}{} \\
&\multicolumn{10}{c}{\textbf{SSIM}} \\
 & \scenename{bicycle} & \scenename{bonsai}  & \scenename{counter}  & \scenename{flowers}  & \scenename{garden}  & \scenename{kitchen}  & \scenename{room} & \scenename{stump} &\scenename{treehill}& \scenename{Average} 
 \\ 
 \hline 
Full Model $\nicefrac{1}{8}$ &0.833 & 0.963 & 0.934 & 0.719 & 0.901 & 0.964 & 0.962 & 0.825 & 0.729 & \textbf{0.870} \\
Full Model $\nicefrac{1}{16}$&0.921 & 0.978 & 0.960 & 0.841 & 0.957 & 0.981 & 0.975 & 0.900 & 0.845 & \textbf{0.929} \\
Full Model $\nicefrac{1}{32}$&0.954 & 0.986 & 0.975 & 0.909 & 0.974 & 0.989 & 0.982 & 0.942 & 0.913 & \textbf{0.958} \\
\hline
w/o LOD $\nicefrac{1}{8}$&0.823 & 0.954 & 0.927 & 0.711 & 0.889 & 0.954 & 0.957 & 0.817 & 0.721 & 0.862 \\
w/o LOD $\nicefrac{1}{16}$&0.917 & 0.974 & 0.958 & 0.839 & 0.952 & 0.974 & 0.973 & 0.897 & 0.841 & 0.925 \\
w/o LOD $\nicefrac{1}{32}$&0.947 & 0.980 & 0.972 & 0.907 & 0.967 & 0.982 & 0.979 & 0.936 & 0.907 & 0.953 \\
\hline
w/o EWA $\nicefrac{1}{8}$&0.832 & 0.962 & 0.932 & 0.713 & 0.900 & 0.962 & 0.959 & 0.826 & 0.733 & 0.869\\
w/o EWA $\nicefrac{1}{16}$&0.919 & 0.978 & 0.958 & 0.836 & 0.956 & 0.980 & 0.972 & 0.897 & 0.844 & 0.927\\
w/o EWA $\nicefrac{1}{32}$&0.945 & 0.983 & 0.970 & 0.897 & 0.971 & 0.985 & 0.978 & 0.925 & 0.902 & 0.951 \\

 \multicolumn{10}{c}{} \\
&\multicolumn{10}{c}{\textbf{LPIPS}} \\
 & \scenename{bicycle} & \scenename{bonsai}  & \scenename{counter}  & \scenename{flowers}  & \scenename{garden}  & \scenename{kitchen}  & \scenename{room} & \scenename{stump} &\scenename{treehill}& \scenename{Average} 
 \\ 
 \hline 

Full Model $\nicefrac{1}{8}$ &0.178 & 0.052 & 0.074 & 0.257 & 0.092 & 0.039 & 0.058 & 0.169 & 0.276 & \textbf{0.133} \\
Full Model $\nicefrac{1}{16}$&0.067 & 0.021 & 0.038 & 0.141 & 0.030 & 0.017 & 0.027 & 0.085 & 0.157 & \textbf{0.065} \\
Full Model $\nicefrac{1}{32}$&0.036 & 0.012 & 0.021 & 0.083 & 0.018 & 0.010 & 0.018 & 0.054 & 0.090 & \textbf{0.038} \\
\hline
w/o LOD $\nicefrac{1}{8}$&0.183 & 0.061 & 0.082 & 0.261 & 0.098 & 0.049 & 0.064 & 0.176 & 0.283 & 0.140 \\
w/o LOD $\nicefrac{1}{16}$&0.071 & 0.027 & 0.041 & 0.141 & 0.035 & 0.029 & 0.030 & 0.089 & 0.161 & 0.069 \\
w/o LOD $\nicefrac{1}{32}$&0.042 & 0.018 & 0.027 & 0.085 & 0.025 & 0.021 & 0.022 & 0.060 & 0.094 & 0.044 \\ 
\hline
w/o EWA $\nicefrac{1}{8}$&0.184 & 0.054 & 0.078 & 0.268 & 0.097 & 0.042 & 0.061 & 0.175 & 0.281 & 0.138\\
w/o EWA $\nicefrac{1}{16}$&0.068 & 0.021 & 0.040 & 0.144 & 0.031 & 0.019 & 0.029 & 0.089 & 0.156 & 0.066\\
w/o EWA $\nicefrac{1}{32}$&0.042 & 0.014 & 0.026 & 0.092 & 0.021 & 0.014 & 0.021 & 0.064 & 0.091 & 0.043\\

\multicolumn{10}{c}{}
    \end{tabular}
    \caption{
    \textbf{Quantitative Results of the Ablation Study on the Mip-NeRF 360 Dataset~\cite{mipnerf360}.} All methods are trained and tested on multiscale inputs. We present the experimental results of our full model alongside methods that exclude either the LOD module or the EWA filter across different scenes and resolutions.
}
    \label{tab:ablation_mip360}
\end{table*}

\begin{table*}[h]
    \centering
    \small
    \begin{tabular}{l|cccccccc|c}
         & \multicolumn{9}{c}{\textbf{PSNR}} \\
 & \scenename{chair}  & \scenename{drums}  & \scenename{ficus}  & \scenename{hotdog}  & \scenename{lego}  & \scenename{materials}  & \scenename{mic}  & \scenename{ship} & \scenename{Average} 
 \\ 
 \hline 
NeRF~\cite{mildenhall2020nerf}&                    33.39  &                    25.87  &                    30.37  &                    35.64  &                    31.65  &                       30.18  &                    32.60  &                    30.09  & 31.23
\\
Mip-NeRF~\cite{mip-nerf}&                   37.14  &                    27.02  &                    33.19  &                    \cellcolor{yellow}39.31  &                   \cellcolor{yellow}35.74  &                    \cellcolor{tablered}32.56  &                   \cellcolor{orange}38.04  &                    \cellcolor{yellow}33.08  & 34.51
\\
\hline
Plenoxels~\cite{fridovich2022plenoxels}&                    32.79  &                    25.25  &                    30.28  &                    34.65  &                    31.26  &                    28.33  &                    31.53  &                    28.59 & 30.34 
\\
TensoRF~\cite{chen2022tensorf}&                    32.47  &                    25.37  &                    31.16  &                    34.96  &                    31.73  &                    28.53  &                    31.48  &                    29.08  & 30.60
\\
Instant-ngp~\cite{muller2022instant}&                    32.95  &                    26.43  &                    30.41  &                    35.87  &                    31.83  &                    29.31  &                    32.58  &                    30.23  & 31.20
\\
Tri-MipRF~\cite{trimip-nerf}*& 37.67 & 27.35 & 33.57 & 38.78 & 35.72 & 31.42 & 37.63 & 32.74 & 34.36
\\
\hline
3DGS~\cite{kerbl3Dgaussians}& 32.73 & 25.30 & 29.00 & 35.03 & 29.44 & 27.13 & 31.17 & 28.33 & 29.77
\\
Mipmap-GS~\cite{mipmapgs} &32.72 & 25.30 & 29.01 & 35.01 & 29.45 & 27.14 & 31.17 & 28.43 & 29.78
\\
MSGS~\cite{multiscale-3dgs}&27.00 & 21.16 & 25.97 & 28.80 & 25.35 & 23.14 & 24.46 & 23.16 & 24.88
\\
Mip-Splatting ~\cite{mip-splatting}&\cellcolor{yellow}37.48 &\cellcolor{yellow} 27.74 &\cellcolor{yellow} 34.71 & 39.15 & 35.07 & \cellcolor{yellow}31.88 & 37.68 & 32.80 & \cellcolor{yellow}34.56
\\
Analytic-Splatting ~\cite{analytic-splatting}&\cellcolor{orange}38.26 &\cellcolor{orange}27.98 &\cellcolor{tablered}36.11 &\cellcolor{orange}39.47 &\cellcolor{orange}35.75 &31.74 &\cellcolor{yellow}37.78 &\cellcolor{orange}33.13 &\cellcolor{orange}35.03 \\
LOD-GS (ours) &\cellcolor{tablered}38.86 &\cellcolor{tablered}28.01 &\cellcolor{orange}35.90 &\cellcolor{tablered}39.86 &\cellcolor{tablered}36.20 &\cellcolor{orange}32.33 &\cellcolor{tablered}38.41 &\cellcolor{tablered}33.38 &\cellcolor{tablered}35.37 \\

\multicolumn{9}{c}{} \\
 & \multicolumn{9}{c}{\textbf{SSIM}} \\
 & \scenename{chair}  & \scenename{drums}  & \scenename{ficus}  & \scenename{hotdog}  & \scenename{lego}  & \scenename{materials}  & \scenename{mic}  & \scenename{ship} & \scenename{Average} 
\\ 
\hline 
NeRF~\cite{mildenhall2020nerf}&                    0.971  &                    0.932  &                    0.971  &                    0.979  &                    0.965  &                       0.967  &                    0.980  &                    0.900 &0.958  
\\
Mip-NeRF~\cite{mip-nerf}&                     0.988  &                     0.945  &                     0.984  &                     0.988  &                     0.984  &                      \cellcolor{orange}0.977  &                      0.993  &                     0.922  &  0.973
\\
\hline
Plenoxels~\cite{fridovich2022plenoxels}&                    0.968  &                    0.929  &                    0.972  &                    0.976  &                    0.964  &                    0.959  &                    0.979  &                    0.892 & 0.955  
\\
TensoRF~\cite{chen2022tensorf,chen2022tensorf}&                    0.967  &                    0.930  &                    0.974  &                    0.977  &                    0.967  &                    0.957  &                    0.978  &                    0.895  & 0.956 
\\
Instant-ngp~\cite{muller2022instant}&                    0.971  &                    0.940  &                    0.973  &                    0.979  &                    0.966  &                    0.959  &                    0.981  &                    0.904  & 0.959
\\
Tri-MipRF~\cite{trimip-nerf}*& 0.990 & 0.951 & 0.985 & 0.988 & 0.986 & 0.969 & 0.992 & 0.929 & 0.974
\\
\hline
3DGS~\cite{kerbl3Dgaussians}& 0.976 & 0.941 & 0.968 & 0.982 & 0.964 & 0.956 & 0.979 & 0.910 & 0.960
\\
Mipmap-GS~\cite{mipmapgs}& 0.976 & 0.941 & 0.968 & 0.982 & 0.964 & 0.956 & 0.979 & 0.911 & 0.960
\\
MSGS~\cite{multiscale-3dgs} &0.915 & 0.849 & 0.920 & 0.929 & 0.884 & 0.883 & 0.910 & 0.828 & 0.890
\\
Mip-Splatting ~\cite{mip-splatting}& \cellcolor{orange}0.991 & \cellcolor{orange}0.963 & \cellcolor{orange}0.990 &\cellcolor{orange} 0.990 & \cellcolor{orange}0.987 & \cellcolor{tablered}0.978 & \cellcolor{tablered}0.994 &\cellcolor{orange} 0.936 &\cellcolor{orange}0.979
\\
Analytic-Splatting ~\cite{analytic-splatting}&\cellcolor{tablered}0.992 &\cellcolor{tablered}0.964 &\cellcolor{tablered}0.992 &\cellcolor{tablered}0.991 &\cellcolor{tablered}0.988 &\cellcolor{orange}0.977 &\cellcolor{tablered}0.994 &\cellcolor{orange}0.936 &\cellcolor{orange}0.979 \\
LOD-GS (ours) &\cellcolor{tablered}0.992 &\cellcolor{tablered}0.964 &\cellcolor{tablered}0.992 &\cellcolor{tablered}0.991 &\cellcolor{tablered}0.988 &\cellcolor{tablered}0.978 &\cellcolor{tablered}0.994 &\cellcolor{tablered}0.938 &\cellcolor{tablered}0.980 \\

\multicolumn{9}{c}{} \\
 & \multicolumn{9}{c}{\textbf{LPIPS}} \\
 & \scenename{chair}  & \scenename{drums}  & \scenename{ficus}  & \scenename{hotdog}  & \scenename{lego}  & \scenename{materials}  & \scenename{mic}  & \scenename{ship} & \scenename{Average} 
\\ 
\hline 
NeRF~\cite{mildenhall2020nerf}&                    0.028  &                       0.059  &                    0.026  &                       0.024  &                    0.035  &                       0.033  &                    0.025  &                       0.085  & 0.044
\\
Mip-NeRF~\cite{mip-nerf}&                     0.011  &                     0.044  &                      0.014  &                      0.012  &                     0.013  &                      0.019  &                      0.007  &                      0.062  &  0.026
\\
\hline
Plenoxels~\cite{fridovich2022plenoxels}&                    0.040  &                    0.070  &                    0.032  &                    0.037  &                    0.038  &                    0.055  &                    0.036  &                    0.104  & 0.051
\\
TensoRF~\cite{chen2022tensorf}&                    0.042  &                    0.075  &                    0.032  &                    0.035  &                    0.036  &                    0.063  &                    0.040  &                    0.112  & 0.054
\\
Instant-ngp~\cite{muller2022instant}&                    0.035  &                    0.066  &                    0.029  &                    0.028  &                    0.040  &                    0.051  &                    0.032  &                    0.095  & 0.047
\\
Tri-MipRF~\cite{trimip-nerf}*&0.011 & 0.046 & 0.016 & 0.014 & 0.013 & 0.033 & 0.008 & 0.069 & 0.026
\\
\hline
3DGS~\cite{kerbl3Dgaussians}& 0.025 & 0.056 & 0.030 & 0.022 & 0.038 & 0.040 & 0.023 & 0.086 & 0.040
\\
Mipmap-GS~\cite{mipmapgs}&0.025 & 0.055 & 0.030 & 0.022 & 0.038 & 0.040 & 0.023 & 0.086 & 0.040
\\
MSGS~\cite{multiscale-3dgs}&0.046 & 0.090 & 0.056 & 0.037 & 0.065 & 0.057 & 0.048 & 0.113 & 0.064
\\
Mip-Splatting ~\cite{mip-splatting}&\cellcolor{yellow} 0.010 & \cellcolor{yellow}0.031 & \cellcolor{yellow}0.009 & \cellcolor{tablered}0.011 & \cellcolor{orange}0.012 & \cellcolor{tablered}0.018 & \cellcolor{tablered}0.005 & \cellcolor{orange}0.059 &\cellcolor{orange} 0.019
\\
Analytic-Splatting ~\cite{analytic-splatting}&\cellcolor{orange}0.008 &\cellcolor{orange}0.029 &\cellcolor{tablered}0.007 &\cellcolor{tablered}0.011 &\cellcolor{tablered}0.011 &\cellcolor{tablered}0.018 &\cellcolor{tablered}0.005 &\cellcolor{tablered}0.058 &\cellcolor{tablered}0.018 \\
LOD-GS (ours) &\cellcolor{tablered}0.007 &\cellcolor{tablered}0.028 &\cellcolor{orange}0.008 &\cellcolor{tablered}0.011 &\cellcolor{tablered}0.011 &\cellcolor{tablered}0.018 &\cellcolor{tablered}0.005 &\cellcolor{tablered}0.058 &\cellcolor{tablered}0.018 \\
\multicolumn{9}{c}{}

    \end{tabular}
    \caption{\textbf{Multi-scale Training and Multi-scale Testing on the the Blender dataset~\cite{mildenhall2020nerf}}. For each scene, we report the arithmetic mean of each metric averaged over the four scales used in the dataset (full resolution, $\nicefrac{1}{2}$, $\nicefrac{1}{4}$, and $\nicefrac{1}{8}$ downsampled scales).
    }
    \label{tab:multiscale_blender_supp}
\end{table*}

\begin{table*}[t]
    \centering
    \small
    \begin{tabular}{l|ccccccc|c}
        & \multicolumn{8}{c}{\textbf{PSNR}} \\
 & \scenename{chair} & \scenename{ficus}  & \scenename{hotdog}  & \scenename{lego}  & \scenename{materials}  & \scenename{mic}  & \scenename{ship} & \scenename{Average} 
 \\ 
 \hline 
3DGS~\cite{kerbl3Dgaussians} &33.71 & 32.28 & 33.10 & 33.14 & 33.36 & 30.64 & 29.86 & 32.30 \\ 
Mimpmap-GS~\cite{mipmapgs} &32.80 & 30.86 & 32.55 & 31.20 & 31.72 & 29.21 & 28.33 & 30.95 \\ 
MSGS~\cite{multiscale-3dgs}&32.78 & 29.86 & 32.56 & 31.19 & 31.76 & 29.33 & 28.37 & 30.84 \\ 
Mip-Splatting\cite{mip-splatting}&\cellcolor{yellow}35.23 & \cellcolor{yellow}33.42 & \cellcolor{yellow}34.09 & \cellcolor{orange}35.06 & \cellcolor{orange}34.93 & \cellcolor{orange}31.00 & \cellcolor{orange}31.27 & \cellcolor{yellow}33.57 \\ 
Analytic-Splatting\cite{analytic-splatting}&\cellcolor{orange}35.63 & \cellcolor{orange}33.52 & \cellcolor{orange}34.26 & \cellcolor{yellow}35.00 & \cellcolor{yellow}34.71 & \cellcolor{yellow}30.72 & \cellcolor{yellow}31.24 & \cellcolor{orange}33.58 \\ 
LOD-GS(ours) &\cellcolor{tablered}35.90 &\cellcolor{tablered}33.94 & \cellcolor{tablered}34.39 & \cellcolor{tablered}35.34 &\cellcolor{tablered} 35.09 &\cellcolor{tablered} 31.58 & \cellcolor{tablered}31.69 & \cellcolor{tablered}33.99 \\ 

% -------------------------------------------------------------------
 \multicolumn{8}{c}{} \\
 & \multicolumn{8}{c}{\textbf{SSIM}} \\
 & \scenename{chair}  & \scenename{ficus}  & \scenename{hotdog}  & \scenename{lego}  & \scenename{materials}  & \scenename{mic}  & \scenename{ship} & \scenename{Average} 
 \\
 \hline
3DGS~\cite{kerbl3Dgaussians} &0.971 & 0.975 & \cellcolor{orange}0.962 & 0.958 & 0.977 & \cellcolor{tablered}0.947 & \cellcolor{tablered}0.910 & \cellcolor{orange}0.957 \\
Mimpmap-GS~\cite{mipmapgs} &0.967 & 0.971 & 0.960 & 0.953 & 0.974 & 0.941 & 0.902 & 0.952  \\ 
MSGS~\cite{multiscale-3dgs}&0.967 & 0.969 & 0.960 & 0.953 & 0.974 & 0.942 & 0.902 & 0.952 \\
Mip-Splatting\cite{mip-splatting}&\cellcolor{orange}0.972 & \cellcolor{orange}0.976 & \cellcolor{orange}0.962 & \cellcolor{orange}0.960 & \cellcolor{tablered}0.979 & \cellcolor{orange}0.943 & \cellcolor{orange}0.909 & \cellcolor{orange}0.957 \\
Analytic-Splatting\cite{analytic-splatting}&\cellcolor{tablered}0.973 & \cellcolor{tablered}0.977 & \cellcolor{orange}0.962 & \cellcolor{orange}0.960 & \cellcolor{yellow}0.977 & 0.939 & 0.908 & \cellcolor{orange}0.957 \\
LOD-GS(ours) &\cellcolor{tablered}0.973 & \cellcolor{tablered}0.977 & \cellcolor{tablered}0.963 & \cellcolor{tablered}0.961 & \cellcolor{orange}0.978 & \cellcolor{tablered}0.947 & \cellcolor{tablered}0.910 & \cellcolor{tablered}0.958 \\

 % -------------------------------------------------------------------
\multicolumn{8}{c}{} \\
 & \multicolumn{8}{c}{\textbf{LPIPS}} \\
 & \scenename{chair}  & \scenename{ficus}  & \scenename{hotdog}  & \scenename{lego}  & \scenename{materials}  & \scenename{mic}  & \scenename{ship} & \scenename{Average} 
\\ 
\hline
3DGS~\cite{kerbl3Dgaussians} &\cellcolor{yellow}0.026 & 0.028 &\cellcolor{orange} 0.111 & 0.063 & 0.041 & \cellcolor{orange}0.056 & \cellcolor{tablered}0.108 & \cellcolor{orange}0.062 \\ 
Mimpmap-GS~\cite{mipmapgs} &0.028 & 0.030 & 0.113 & 0.068 & 0.043 & 0.061 & 0.115 & 0.066 \\ 
MSGS~\cite{multiscale-3dgs}&0.028 & 0.032 & 0.113 & 0.068 & 0.043 & 0.060 & 0.115 & 0.066 \\ 
Mip-Splatting\cite{mip-splatting}&\cellcolor{yellow}0.026 &\cellcolor{yellow} 0.027 & 0.114 & \cellcolor{orange}0.062 & \cellcolor{tablered}0.039 & \cellcolor{yellow}0.058 & \cellcolor{orange}0.109 & \cellcolor{orange}0.062 \\ 
Analytic-Splatting\cite{analytic-splatting}&\cellcolor{tablered}0.024 & \cellcolor{orange}0.026 & \cellcolor{orange}0.111 & \cellcolor{orange}0.061 & \cellcolor{yellow}0.041 & 0.062 & 0.111 & \cellcolor{orange}0.062 \\ 
LOD-GS(ours) &\cellcolor{orange}0.025 & \cellcolor{tablered}0.025 & \cellcolor{tablered}0.109 & \cellcolor{tablered}0.059 & \cellcolor{orange}0.040 & \cellcolor{tablered}0.055 & \cellcolor{tablered}0.108 & \cellcolor{tablered}0.060\\
\multicolumn{8}{c}{}
    \end{tabular}
    \vspace{-1em}
    \caption{\textbf{Multi-Level Training and Multi-Level Testing on our extended Blender dataset~\cite{mildenhall2020nerf}}. For each scene, we report the arithmetic mean of each metric averaged over the 3 Levels used in the dataset (near, middle and far). 
    }
    \label{tab:multilevel_blender_supp}
\end{table*}

\begin{table*}[t]
    \centering
    \small
    \begin{tabular}{l|ccccccccc|c}
        & \multicolumn{10}{c}{\textbf{PSNR}} \\
 & \scenename{bicycle} & \scenename{bonsai}  & \scenename{counter}  & \scenename{flowers}  & \scenename{garden}  & \scenename{kitchen}  & \scenename{room} & \scenename{stump} &\scenename{treehill}& \scenename{Average} 
 \\ 
 \hline 
3DGS~\cite{kerbl3Dgaussians}&27.74 & 31.94 & 29.95 & 25.12 & \cellcolor{yellow}30.05 & 31.51 & 32.23 & 28.57 & 25.62 & 29.19 \\
Mipmap-GS~\cite{mipmapgs} &26.99 & 30.25 & 28.75 & 24.28 & 29.45 & 30.84 & 31.19 & 27.95 & 25.32 & 28.33 \\
MSGS~\cite{multiscale-3dgs}&26.93 & 30.32 & 28.75 & 24.31 & 29.42 & 30.67 & 31.30 & 27.97 & 25.47 & 28.35 \\
Mip-Splatting\cite{mip-splatting}&\cellcolor{yellow}28.65 & \cellcolor{yellow}32.69 & \cellcolor{yellow}30.50 & \cellcolor{yellow}25.86 & 29.93 & \cellcolor{yellow}32.55 & \cellcolor{yellow}33.06 & \cellcolor{yellow}28.96 & \cellcolor{tablered}26.48 & \cellcolor{yellow}29.86 \\
Analytic-Splatting\cite{analytic-splatting} &\cellcolor{orange}28.76 & \cellcolor{orange}33.52 & \cellcolor{orange}31.11 & \cellcolor{orange}26.37 & \cellcolor{orange}30.81 & \cellcolor{orange}33.78 & \cellcolor{orange}33.41 & \cellcolor{orange}29.56 & \cellcolor{yellow}26.01 & \cellcolor{orange}30.37 \\
LOD-GS(ours) &\cellcolor{tablered}29.14 & \cellcolor{tablered}34.34 & \cellcolor{tablered}31.60 & \cellcolor{tablered}26.65 & \cellcolor{tablered}31.32 & \cellcolor{tablered}34.23 & \cellcolor{tablered}33.85 & \cellcolor{tablered}29.88 & \cellcolor{orange}26.19 & \cellcolor{tablered}30.80 \\

 \multicolumn{10}{c}{} \\
&\multicolumn{10}{c}{\textbf{SSIM}} \\
 & \scenename{bicycle} & \scenename{bonsai}  & \scenename{counter}  & \scenename{flowers}  & \scenename{garden}  & \scenename{kitchen}  & \scenename{room} & \scenename{stump} &\scenename{treehill}& \scenename{Average} 
 \\ 
 \hline 
3DGS~\cite{kerbl3Dgaussians}&0.877 & 0.964 & 0.943 & 0.795 & 0.928 & 0.964 & 0.965 & 0.862 & 0.811 & 0.901 \\
Mipmap-GS~\cite{mipmapgs} &0.860 & 0.952 & 0.930 & 0.770 & 0.921 & 0.961 & 0.955 & 0.842 & 0.803 & 0.888 \\
MSGS~\cite{multiscale-3dgs}&0.860 & 0.953 & 0.930 & 0.769 & 0.921 & 0.960 & 0.957 & 0.842 & 0.805 & 0.888 \\
Mip-Splatting\cite{mip-splatting}&\cellcolor{orange}0.899 & \cellcolor{yellow}0.971 & \cellcolor{yellow}0.951 & \cellcolor{yellow}0.816 & \cellcolor{yellow}0.938 & \cellcolor{yellow}0.966 & \cellcolor{orange}0.970 & \cellcolor{yellow}0.874 & \cellcolor{yellow}0.831 & \cellcolor{yellow}0.913 \\
Analytic-Splatting\cite{analytic-splatting} &\cellcolor{yellow}0.897 & \cellcolor{orange}0.972 & \cellcolor{orange}0.954 & \cellcolor{orange}0.821 & \cellcolor{orange}0.940 & \cellcolor{orange}0.975 & \cellcolor{orange}0.970 & \cellcolor{orange}0.884 & \cellcolor{orange}0.825 & \cellcolor{orange}0.915 \\
LOD-GS(ours) &\cellcolor{tablered}0.902 & \cellcolor{tablered}0.976 & \cellcolor{tablered}0.956 & \cellcolor{tablered}0.824 & \cellcolor{tablered}0.944 & \cellcolor{tablered}0.979 &\cellcolor{tablered} 0.973 &\cellcolor{tablered} 0.888 & \cellcolor{tablered}0.829 & \cellcolor{tablered}0.919 \\

 \multicolumn{10}{c}{} \\
&\multicolumn{10}{c}{\textbf{LPIPS}} \\
 & \scenename{bicycle} & \scenename{bonsai}  & \scenename{counter}  & \scenename{flowers}  & \scenename{garden}  & \scenename{kitchen}  & \scenename{room} & \scenename{stump} &\scenename{treehill}& \scenename{Average} 
 \\ 
 \hline 
3DGS~\cite{kerbl3Dgaussians}&0.115 & 0.043 & 0.060 & 0.183 & 0.062 & \cellcolor{yellow}0.037 & 0.045 & 0.130 & 0.192 & 0.096 \\
Mipmap-GS~\cite{mipmapgs} &0.131 & 0.053 & 0.072 & 0.204 & 0.069 & 0.039 & 0.056 & 0.150 & 0.199 & 0.108 \\
MSGS~\cite{multiscale-3dgs}&0.131 & 0.052 & 0.072 & 0.204 & 0.069 & 0.039 & 0.054 & 0.150 & 0.200 & 0.108 \\
Mip-Splatting\cite{mip-splatting}&\cellcolor{tablered}0.090 & \cellcolor{orange}0.030 & \cellcolor{yellow}0.049 & \cellcolor{tablered}0.146 & \cellcolor{yellow}0.053 & \cellcolor{yellow}0.037 & \cellcolor{orange}0.036 & \cellcolor{orange}0.119 & \cellcolor{tablered}0.156 & \cellcolor{orange}0.080 \\
Analytic-Splatting\cite{analytic-splatting} &\cellcolor{orange}0.093 & \cellcolor{yellow}0.032 & \cellcolor{orange}0.046 & \cellcolor{orange}0.157 & \cellcolor{orange}0.048 & \cellcolor{orange}0.027 & \cellcolor{yellow}0.038 & \cellcolor{tablered}0.103 & \cellcolor{orange}0.170 & \cellcolor{orange}0.079 \\
LOD-GS(ours) &\cellcolor{orange}0.093 & \cellcolor{tablered}0.028 & \cellcolor{tablered}0.044 & \cellcolor{yellow}0.160 & \cellcolor{tablered}0.047 & \cellcolor{tablered}0.021 & \cellcolor{tablered}0.034 & \cellcolor{tablered}0.103 & \cellcolor{yellow}0.173 & \cellcolor{tablered}0.078 \\ \multicolumn{10}{c}{}
    \end{tabular}
    \vspace{-1em}
    \caption{\textbf{Multi-Scale Training and Multi-Scale Testing on the the Mip-NeRF 360 dataset~\cite{mipnerf360}}. For each scene, we report the arithmetic mean of each metric averaged over the 3 scales used in the dataset ($\nicefrac{1}{8}$, $\nicefrac{1}{16}$, and $\nicefrac{1}{32}$ downsampled scales). 
    }
    \label{tab:multiscale_360_supp}

\end{table*}

\end{document}